\RequirePackage{fix-cm}

\documentclass[twocolumn]{svjour3}

\smartqed

\usepackage{times}
\usepackage{natbib}
\usepackage{tabularx}
\usepackage[xcdraw,table]{xcolor}
\usepackage[pdftex]{graphicx}
\usepackage{fancybox}
\usepackage[export]{adjustbox}
\graphicspath{{./}}
\DeclareGraphicsExtensions{.pdf,.png}
\usepackage{mathptmx}
\usepackage[cmex10]{amsmath}
\usepackage{amssymb}
\usepackage{algorithm}
\usepackage{booktabs}
\usepackage{lmodern}
\usepackage{pgfplots}
\usepackage{pgfplotstable}
\usetikzlibrary{pgfplots.groupplots}
\usepgfplotslibrary{groupplots}
\pgfplotsset{compat=1.9}
\usepackage{interval}
\usepackage{bm}
\usepackage{xspace}
\usepackage{xfrac}
\usepackage{tikz}
\usetikzlibrary{positioning}
\usetikzlibrary{shapes}
\usetikzlibrary{arrows}
\usepackage{algpseudocode}
\usepackage{footmisc}

\usepackage[labelsep=space,font=small,labelfont=bf]{caption}
\usepackage{subcaption}

\usepackage{units}
\usepackage{threeparttable}
\usepackage{dcolumn}
\usepackage{paralist}
\usepackage{multirow}
\usepackage{wasysym}
\usepackage{url}
\usepackage[threshold=0]{csquotes}
\SetBlockEnvironment{quotation}
\newcommand\MYhyperrefoptions{bookmarks=true,bookmarksnumbered=true,
    pdfpagemode={UseOutlines},plainpages=false,pdfpagelabels=true,
    colorlinks=true,linkcolor={black},citecolor={black},urlcolor={black},
    pdftitle={Online Mutual Foreground Segmentation for Multispectral Stereo Videos},
    pdfsubject={Journal},
    pdfauthor={Pierre-Luc St-Charles}}
\usepackage[\MYhyperrefoptions,pdftex]{hyperref}

\definecolor{lightgray}{gray}{0.8}
\definecolor{bitgray}{gray}{0.6}
\definecolor{darkred}{rgb}{0.85,0,0}
\definecolor{darkgreen}{rgb}{0,0.7,0}
\definecolor{darkblue}{rgb}{0,0,0.6}

\newcolumntype{M}[1]{>{\centering\arraybackslash}m{#1}}
\newcolumntype{N}{@{}m{0pt}@{}}
\newcommand{\specialcell}[2][c]{\begin{tabular}[#1]{@{}c@{}}#2\end{tabular}}

\newcommand{\xor}{\mathbin{\oplus}}
\newcommand{\smallminus}{\scalebox{0.75}[1.0]{\(-\)}}
\newcommand{\smallplus}{\scalebox{0.85}[1.0]{\(+\)}}
\newcommand{\smallequal}{\scalebox{0.85}[1.0]{\(=\)}}

\newcolumntype{C}{>{\centering\arraybackslash}X}

\newcommand{\Xagr}[1]{\mathcal{#1}}
\DeclareMathOperator*{\argmin}{argmin}

\DeclareMathOperator*{\mmax}{max}

\newcommand{\figResultsRowWideHeader}[2]{
    \multicolumn{2}{c}{\bf{#1}} & \multicolumn{2}{c}{\bf{#2}}\\
}

\newcommand{\figResultsRowHeader}[4]{
    \small{#1} & \small{#2} & \small{#3} & \small{#4} \vspace{-.25mm}\\
}

\newcommand{\figResultsRow}[4]{
    \includegraphics[width=.24\linewidth,frame]{#1}\hspace{-.5mm} &
    \includegraphics[width=.24\linewidth,frame]{#2}\hspace{.5mm} &
    \includegraphics[width=.24\linewidth,frame]{#3}\hspace{-.5mm} &
    \includegraphics[width=.24\linewidth,frame]{#4}\vspace{-.5mm}\\
}

\journalname{Journal}

\begin{document}

\title{Online Mutual Foreground Segmentation for Multispectral Stereo Videos}
\author{%
    Pierre-Luc~St-Charles \and
    Guillaume-Alexandre~Bilodeau \and
    Robert~Bergevin%
}

\institute{
    P.-L. St-Charles \at
        Polytechnique Montr{\'e}al \\
        2900, boul. {\'E}douard-Montpetit, Mont{r\'e}al, QC, Canada \\
        \email{pierre-luc.st-charles@polymtl.ca}
    \and
    G.-A. Bilodeau \at
        Polytechnique Montr{\'e}al \\
        2900, boul. {\'E}douard-Montpetit, Mont{r\'e}al, QC, Canada \\
        \email{guillaume-alexandre.bilodeau@polymtl.ca}
    \and
    R. Bergevin \at
        Universit{\'e} Laval \\
        2325, rue de l'Universit{\'e}, Qu{\'e}bec, QC, Canada \\
        \email{robert.bergevin@gel.ulaval.ca}
}

\date{Received: XXXX XX, 2018 / Accepted: XXXX XX, 2018}

\maketitle

\begin{abstract}

The segmentation of video sequences into foreground and background regions is a low-level process commonly used in video content analysis and smart surveillance applications. Using a multispectral camera setup can improve this process by providing more diverse data to help identify objects despite adverse imaging conditions. The registration of several data sources is however not trivial if the appearance of objects produced by each sensor differs substantially. This problem is further complicated when parallax effects cannot be ignored when using close-range stereo pairs. In this work, we present a new method to simultaneously tackle multispectral segmentation and stereo registration. Using an iterative procedure, we estimate the labeling result for one problem using the provisional result of the other. Our approach is based on the alternating minimization of two energy functions that are linked through the use of dynamic priors. We rely on the integration of shape and appearance cues to find proper multispectral correspondences, and to properly segment objects in low contrast regions. We also formulate our model as a frame processing pipeline using higher order terms to improve the temporal coherence of our results. Our method is evaluated under different configurations on multiple multispectral datasets, and our implementation is available online.


\end{abstract}

\section{Introduction}
\label{sec:intro}

The detection and segmentation of objects of interest based on motion analysis in video sequences is a fundamental early vision task. In the context of video surveillance and intelligent environments, objects of interest (or ``foreground'' objects) are disruptors that temporarily break the natural state of the observed scene (the ``background''). Several types of approaches exist to classify image regions as being ``of interest'' based on this criteria \citep[see][]{bouwmans2014_csr,perazzi2016}. While these all have different qualities, they suffer from the same fundamental drawback: if the contrast between an observed object and the background becomes too low, our ability to detect and segment it automatically deteriorates. This problem is not specific to the visible light spectrum, as this camouflaging can occur with any imaging modality.

\begin{figure}[t]
    \centering
    \includegraphics[width=\linewidth]{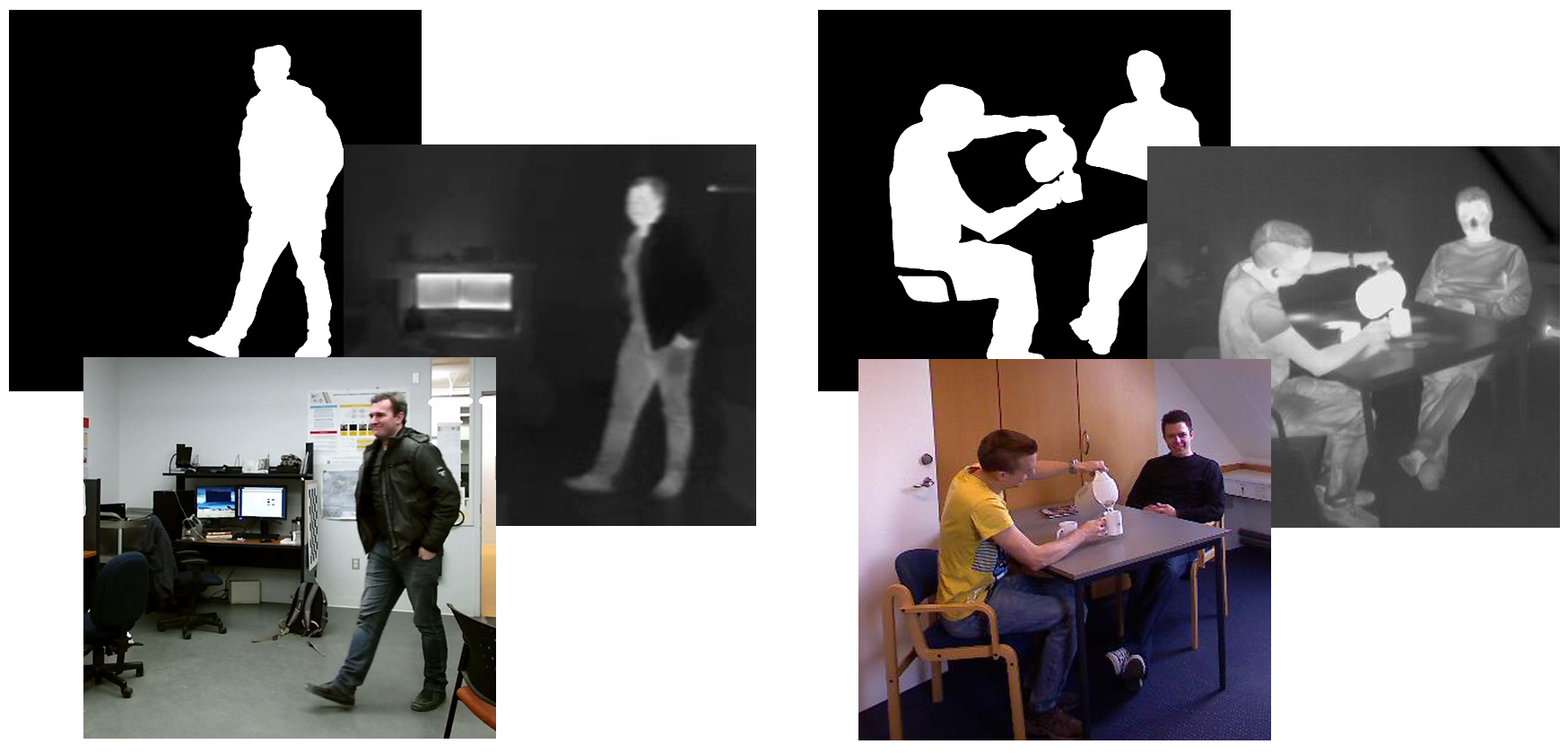}
    \vspace{-12pt}
    \caption{Examples of mutual foreground segmentation in low contrast conditions for RGB-LWIR image pairs. On the left, the person is only partly perceptible in the LWIR spectrum due to a winter coat, but is clearly perceptible in the visible spectrum. The opposite is true on the right, where legs are hard to perceive in the visible spectrum, but easy to perceive in the LWIR spectrum.}
    \label{fig:rgb_lwir_low_contrast_ex}
\end{figure}

However, interestingly, the phenomena describing the appearance of an object and the conditions under which it becomes harder to identify are rarely shared across several imaging modalities. This is especially true when considering for example the visible and Long-Wavelength Infrared (LWIR) spectra, as the correlation between the temperature of an object and its visible appearance is very weak \citep[see][]{bilodeau2011a}. We show an example of this in Figure~\ref{fig:rgb_lwir_low_contrast_ex}. In fact, many surveillance systems rely on the complementarity of these two imaging modalities to detect abnormal events: the visible spectrum can easily identify large objects near ambient temperatures (e.g. vehicles), and the LWIR spectrum can easily identify objects that exhibit abnormal temperatures (e.g. animals, engine parts).

Integrating data captured from different spectral bands to attain benefits in recognition tasks is however not trivial. If the optical axes of the sensors are not already aligned using a beam splitter, a registration method has to be used to bring data points back into a common coordinate system. The image registration problem has been thoroughly studied for identical sensor pairs, but multispectral registration is fundamentally more challenging \citep[c.f.][]{zitova2003}. Since the appearance of objects cannot be directly relied upon to find local correspondences, higher level image features such as edges have to be used instead. These are typically harder to compute, and often result in a loss of registration accuracy at the pixel level when parallax effects are not negligible.

Past research has focused mostly on the problems of binary (or foreground-background) segmentation and multispectral image fusion/registration as separate issues. Yet, holistic approaches such as the ones of \cite{torabi2012,zhao2014} can outperform combinations of distinct methods on identical tasks. These holistic approaches first optimize registration using foreground object contours or trajectories as high-level features, and then use integrated image data to improve their segmentation. Solving both problems at once would be more beneficial, but this goal implies a ``\emph{chicken-and-egg}'' dilemma: the result of one task is needed to obtain the other. An ideal holistic method should thus adopt an iterative optimization approach to resolve this issue. In the case of video sequences, proposed solutions should also consider the temporal redundancy of data to improve their performance. Finally, in the context of surveillance applications, the entire process should function without any human supervision, and allow frame pairs to be processed one at a time.

In this paper, we propose a holistic method to address both segmentation and registration problems by inferring their solutions alternately using move-making algorithms on a set of conditional random fields. We use self-similarity descriptors and shape cues to find proper pixel-level matches across imaging modalities in non-planar scenes, and integrate image data to improve foreground-background partitioning. This integration is achieved by iteratively refining local color models and shape contour positions while continuously realigning data sources. Our two goals are formulated as distinct energy minimization problems, and we use provisional inference results as dynamic priors to converge to a global solution. We also rely on dynamic temporal connections updated via motion cues to improve segmentation coherence over long image sequences.

Our principled bottom-up approach requires no human intervention, and relies on no prior knowledge of the foreground objects' nature. Our models are formulated so that imaging modalities can be combined without assumptions about their specific characteristics, as image regions containing discriminative data are automatically identified. This power of discrimination is exploited to scale the importance of each imaging modality when registering and integrating pixel-level data. It is also used to speed up shape contour evolution in low contrast regions by reducing penalties for label discontinuities when the other view possesses strong intensity gradients in its corresponding regions. Besides, we tackle foreground-background segmentation in the general case of video surveillance, meaning we assume the scene might contain multiple foreground objects at different depths and scales, and that they might not always be moving. This differs significantly from traditional cosegmentation methods, as we make no assumption regarding the distribution of foreground and background regions in the observed scene.

Through our experiments, we show that our primary goal, mutual foreground segmentation, can be achieved efficiently despite low contrast and other adverse conditions in both visible and LWIR images. Performance evaluations show that our approach outperforms both supervised and unsupervised monocular segmentation methods in terms of $F_1$ score on the VAP dataset of \cite{palmero2016}. Compared to the recent video segmentation method of~\cite{pawcs2016}, our method improves its average $F_1$ score by 13\%, from 0.766 to 0.866. To help future benchmarking on this task, we offer a new multispectral video dataset for the simultaneous evaluation of registration and segmentation performance\footnote{\url{http://www.polymtl.ca/litiv/vid/index.php}}. Finally, we also offer our source code and testing framework online\footnote{\url{https://github.com/plstcharles/litiv}}.

Note that our method was previously introduced \citep{stcharles2017}. Here, beyond presenting an extended description of our approach, we introduce a new spatiotemporal term to our model and study its effect on segmentation accuracy, we conduct an ablation study and test the sensitivity of our main parameters individually, we present new experiments on two pre-existing datasets, we introduce a new non-planar RGB-LWIR video dataset, and we provide a benchmark for the evaluation of segmentation and stereo registration on this new dataset. Our source code and annotations have been made available online for future works tackling a similar problem.

The paper is organized as follows. In Section~\ref{sec:relwork}, we present previous works related to our multispectral mutual segmentation problem, and highlight major differences. In Section~\ref{sec:approach}, we describe our dual modeling approach, inference strategies, and implementation details. In Section~\ref{sec:experiments}, we present parameter and configuration studies, and evaluation results on three publicly available datasets. Lastly, we conclude with some remarks in Section~\ref{sec:conclusion}.

\section{Previous Work}
\label{sec:relwork}


The problem of foreground-background segmentation in images is difficult to tackle without some assumptions or constraints. Monocular segmentation solutions typically rely on visual saliency hypotheses (e.g. single foreground object roughly focused) or human supervision to obtain good results \citep{arbelaez2011,rother2004}. The same problem in the temporal domain (i.e. on image sequences) is easier to address due to the additional assumptions that can be made regarding object or scene motion.

Multiple families of methods exist in video segmentation; the main ones are listed here. Background subtraction methods work by building a model representing the background under the assumption that the camera is static. These methods then perform one-class pixel classification to label all outliers as foreground without supervision \citep{bouwmans2014_csr}. These methods are favored in cases where foreground objects can temporarily become immobile, as they will retain their labeling for some time. Other video object segmentation approaches instead extend the concept of visual saliency into the temporal domain using highly connected graph structures \citep{perazzi2016}. These approaches can usually be applied to sequences with changing viewpoints, but are computationally more demanding. Finally, motion clustering methods exist that rely on optical flow or trajectory points partitioning to identify image regions that behave differently from their surroundings \citep{tron2007}. The strong link between motion partitioning and video object segmentation has also become a focus in recent years \citep{jain2017,cheng2017}. Also, in semi-supervised settings, approaches based on end-to-end neural networks have also become increasingly popular for single object video segmentation \citep{cheng2017,caelles2017}.

Foreground-background segmentation can become easier if multiple images of the object(s) of interest are available. Two families of methods have been developed for this circumstance: cosegmentation methods and mutual segmentation methods. Cosegmentation methods typically rely on visual saliency assumptions (e.g. shared foreground appearance and low background correlation across different views), and assume a single object is targeted and shared throughout all views \citep{rother2006,zhu2016}. Interestingly, cosegmentation methods can also work with different object instances from the same object category \citep{vicente2011}. On the other hand, mutual segmentation methods typically assume that the same object instance is observed from multiple viewpoints, and optimize the geometric consistency of the extracted foreground region \citep{djelouah2015,jeong2017,riklin2008}. Our work falls into this second family of methods, as we assume the use of a synchronized stereo pair for data capture.

Previous mutual segmentation methods have typically focused on single-spectrum imaging \citep{riklin2008,ju2015,bleyer2011}, or have used depth sensors to solve the registration problem and to provide a range-based solution for foreground object detection \citep{jeong2017,djelouah2015,zhang2016}. Of these, our proposed method is closest to the work of \cite{riklin2008}, who termed the idea of ``mutual segmentation'' for objects in visible image pairs. Their approach addresses the uncertainty of object boundary localization under occlusions and noise by iteratively optimizing active contours without supervision. Their use of a biased shape term however entails that a free parameter directly controls the elimination of ambiguous shape segments in the image pair. In our work, we avoid this parameterization issue by relying on local saliency and self-refining color models to automatically integrate multiple view data. Our object contours then expand and contract until they naturally converge. Besides, the method of \cite{riklin2008} considers that all images are related only by planar projective homographies, and thus it cannot handle parallax issues in 3D scenes. This latter problem was addressed by \cite{ju2015}, who also proposed a contour-based modeling approach for mutual foreground segmentation in stereo pairs. This more recent approach however relies on the assumption that near-perfect foreground contours obtained via human supervision are available in at least one of the views. Lastly, the work of \cite{bleyer2011} is also somewhat related to ours: they tackle disparity (or parallax) estimation for calibrated stereo pairs using a piecewise planar model based on object segmentation. However, their main goal is scene-wide data registration, which is very computationally demanding. According to \cite{tippetts2016}, processing an image pair took the method about 20 minutes. In our case, we only focus on the registration and segmentation of foreground objects classified as such in a video surveillance mindset. This makes our proposed approach much more lightweight and applicable to real data streams.

\begin{figure*}[t]
    \centering
    \includegraphics[width=0.7\linewidth]{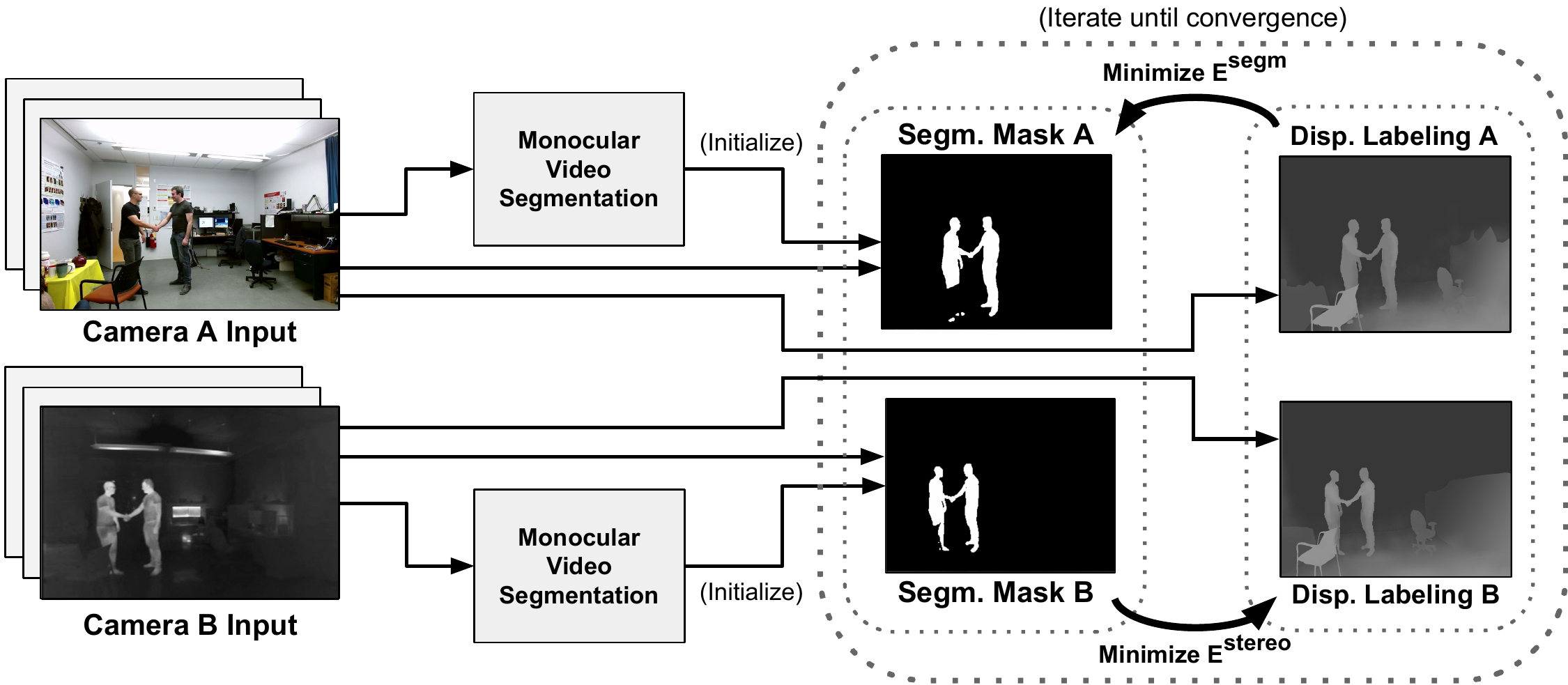}
    \vspace{-6pt}
    \caption{Flowchart of the proposed method. A monocular video segmentation method is first used to initialize segmentation masks for both cameras individually. Then, the energies of the stereo and segmentation models (described in Sections~\ref{sec:approach:stereo} and~\ref{sec:approach:segm}, respectively) are alternately minimized until a proper global solution is reached. The output of our method then consists of the refined segmentation masks of the input frames, and of the reciprocal disparity labelings computed for both cameras.}
    \label{fig:overview}
\end{figure*}

The use of multispectral data (other than RGBD) has been mostly neglected in the context of mutual segmentation or cosegmentation due to the registration problem. As stated before, this difficulty is due to the (typically) low correlation between the appearances of objects in different spectral bands \citep[see][]{zitova2003}. Beam splitters can be used to avoid the registration problem altogether \citep{bienkowski2012,hwang2015}. These setups are however very delicate, and they induce color distortions. Moreover, the elimination of parallax also prevents the recovery of depth information from the scene.

In practice, if the chosen spectral bands are not too distant in terms of their imaging characteristics (e.g. visible light and near-infrared), modern image descriptors and similarity measures can be used to find local correspondences with varying degrees of success \citep[see][]{pinggera2012}. These ``close'' spectrum pairs are however less interesting to integrate in machine vision systems due to their resemblance. On the other hand, traditional appearance-based matching approaches suffer when distant spectrum pairs are selected; see for example the study done for visible (RGB) and Long-Wavelength Infrared (LWIR) pairs by \cite{bilodeau2014}. Multispectral registration thus has to rely on higher level features that encapsulate raw object appearance in order to find proper local correspondences. In the recent literature, some have relied on edge matching in local neighborhoods \citep{coiras2000,mouats2013} or in Hough space \citep{pistarelli2013} to resolve this problem. Edge-based approaches are however more suited to man-made environments, and underperform in more general settings (e.g. open terrain) where large intensity gradients are rarer or more weakly correlated between imaging modalities.

Other works have instead addressed the registration problem in the temporal domain by adopting motion-based cues \citep{torabi2012,zhao2014,nguyen2016}, which is more similar to our approach. In the work of \cite{torabi2012}, the trajectories of foreground objects are used for high-level registration based on the idea that position and motion are fully independent of appearance. In the works of \cite{zhao2014} and \cite{nguyen2016}, foreground shapes obtained via background subtraction are used for contour matching. This latter strategy has been shown to be more pixel-accurate for the registration of foreground objects, but it still depends strongly on the performance of the segmentation method used. In our proposed method, we address this problem by combining contour-based registration and segmentation into a global optimization framework.

Finally, as for the combination of multispectral registration and segmentation, we can highlight the existence of a few papers. \cite{torabi2012} propose a solution based on object-wise planar registration, and improve segmentation masks obtained via background subtraction by combining multispectral data using a sum-rule approach. \cite{zhao2014} also rely on object-wise planar registration, and use multiple object trackers to improve the results of parallel segmentors \emph{a posteriori}. In this case, the methods are run in cascade to resolve the ``\emph{chicken-and-egg}'' optimization dilemma stated earlier. The strategies of \cite{torabi2012} and \cite{zhao2014} do not handle occlusions well due to their high level registration approach, and only provide a single-pass improvement to the segmentation results of a given frame pair. \cite{palmero2016} introduced a human body segmentation method for trimodal (RGBD-LWIR) image sequences based on feature fusion using a random forest classifier. They also avoid pixel-level registration by predefining a set of homographies to use at runtime based on detected foreground object depth. \cite{davis2007} proposed a dual background subtraction model and contour extraction technique to improve RGB-LWIR foreground fusion based on local visual saliency evaluation. Similarly, \cite{li2017} proposed a background subtraction method based on the low-rank decomposition of integrated RGB-LWIR pairs to improve foreground segmentation in a global framework. The main shortcoming of these latter two works is that they only handle planar scenes (i.e. scenes where parallax issues are negligible) using a single predefined homography. To the best of our knowledge, no method has previously been proposed to tackle multispectral non-planar registration and mutual foreground segmentation simultaneously.

\section{Proposed Approach}
\label{sec:approach}

Our approach can be described based on its two main components: the stereo matching model for disparity (or parallax) estimation on epipolar lines, described in Section~\ref{sec:approach:stereo}, and the shape matching model for binary image segmentation, described in Section~\ref{sec:approach:segm}. These two models are conditional random fields formulated as discrete energy functions that tackle the multispectral registration and segmentation problems in an integrated fashion. Our energy functions are minimized alternately using move-making algorithms, as described in Section~\ref{sec:approach:inference}. The flowchart in Figure~\ref{fig:overview} illustrates our approach.

We begin with an introduction of the general terms and notation used in this section. Given a set of rectified images $\Xagr{I}\,{=}\left\{I_k \right\}$ (with $k\,{=}\left\{0,1\right\}$ in the case of a stereo pair), the disparity label space $\Xagr{L}_D{=}\left\{0,\ldots,d_\text{max}\right\}$, and the background-foreground label space $\Xagr{L}_S{=}\left\{0,1\right\}$, our goal is to find the optimal pixel-wise disparity and segmentation labelings $\Xagr{D}\,{=}\left\{\Xagr{D}_k\right\}$ and $\Xagr{S}\,{=}\left\{\Xagr{S}_k\right\}$ such that:
\begin{equation} 
    \Xagr{D}_k = \argmin_{D_k} E^{\text{stereo}}_k\big(D_k\big)
    \label{eq:argmin_stereo},
\end{equation}
\begin{equation} 
    \Xagr{S}_k = \argmin_{S_k} E^{\text{segm}}_k\big(S_k\big)
    \label{eq:argmin_segm},
\end{equation}
where $D_k{=}\left\{d_{p} : p \in I_k, d_{p} \in \Xagr{L}_D \right\}$ is a disparity labeling, $S_k\,{=}\left\{s_{p} : p \in I_k, s_{p} \in \Xagr{L}_S \right\}$ is a segmentation labeling (or mask), and where the energy cost functions $E^{\text{stereo}}_k$ and $E^{\text{segm}}_k$ are described in Sections~\ref{sec:approach:stereo} and~\ref{sec:approach:segm}, respectively. For now, note that these functions are linked through their estimation results, $\Xagr{D}_k$ and $\Xagr{S}_k$, which are used as dynamic priors throughout the minimization. In other words, disparity labels $d_{p}$ for each pixel $p$ in $I_k$ are used in $E^{\text{segm}}_k$ for appearance data integration, and segmentation labels $s_{p}$ are used in $E^{\text{stereo}}_k$ to improve stereo matching. Lastly, note that we sometimes omit the $k$ subscript in the following subsections to simplify the notation, as most equations only deal with one image of the stereo pair at a time.

\subsection{Stereo Registration Model}
\label{sec:approach:stereo}

We tackle the multispectral stereo registration problem for non-planar scenes using a sliding window strategy for pixel matching. This search for correspondences is limited to an horizontal axis on the image plane due to epipolar geometry constraints. These constraints restrict the disparity (or parallax) between the 2D projections of an observed 3D object point to one dimension \citep[see][]{hartley2003}. In short, given the intrinsic and extrinsic parameters of the stereo pair obtained via calibration, we can rectify the input images. This forces the corresponding projection of a 2D point in one view to be located somewhere on the same horizontal line in the other view. While calibration does require human intervention, it is a one-time effort generally accepted in an unsupervised system. It could also be replaced by an automatic approach \cite[e.g.][]{nguyen2016}.

For a pixel-wise disparity label map $D$, we define its energy (or cost) to be minimized as
\begin{equation}
    \begin{aligned}
        E^{\text{stereo}}(D) ={} & E^{\text{appearance}}(D)\,+ E^{\text{shape}}(D) \\
                                 & \,+ E^{\text{uniqueness}}(D)\,+ E^{\text{smooth1}}(D)
    \end{aligned}
    \label{eq:stereo_e_global}.
\end{equation}
Each term in this cost function is crafted to promote a desired property of the output disparity labeling, and is described in detail in the following paragraphs. The first three terms are unary costs summed over all pixels of the image. The appearance and shape terms evaluate the local affinity between a pixel $p$ and its corresponding pixel shifted by $d_p$ in the other view. The uniqueness term penalizes multiple matches with $p$ in the other view. The last term is a sum of pairwise smoothness costs used to penalize irregular disparities in uniform image regions. Note that in order to maximize processing speed for image pair sequences, we keep our stereo model simple. Our results would undoubtedly improve with second-order terms such as those of \cite{woodford2009} or \cite{kohli2009}, but at an important increase in computational complexity. Moreover, since we only focus on the registration of foreground objects, higher-order surface smoothness priors are not as important here.

\begin{figure}[t]
    \centering
    \includegraphics[width=0.95\linewidth]{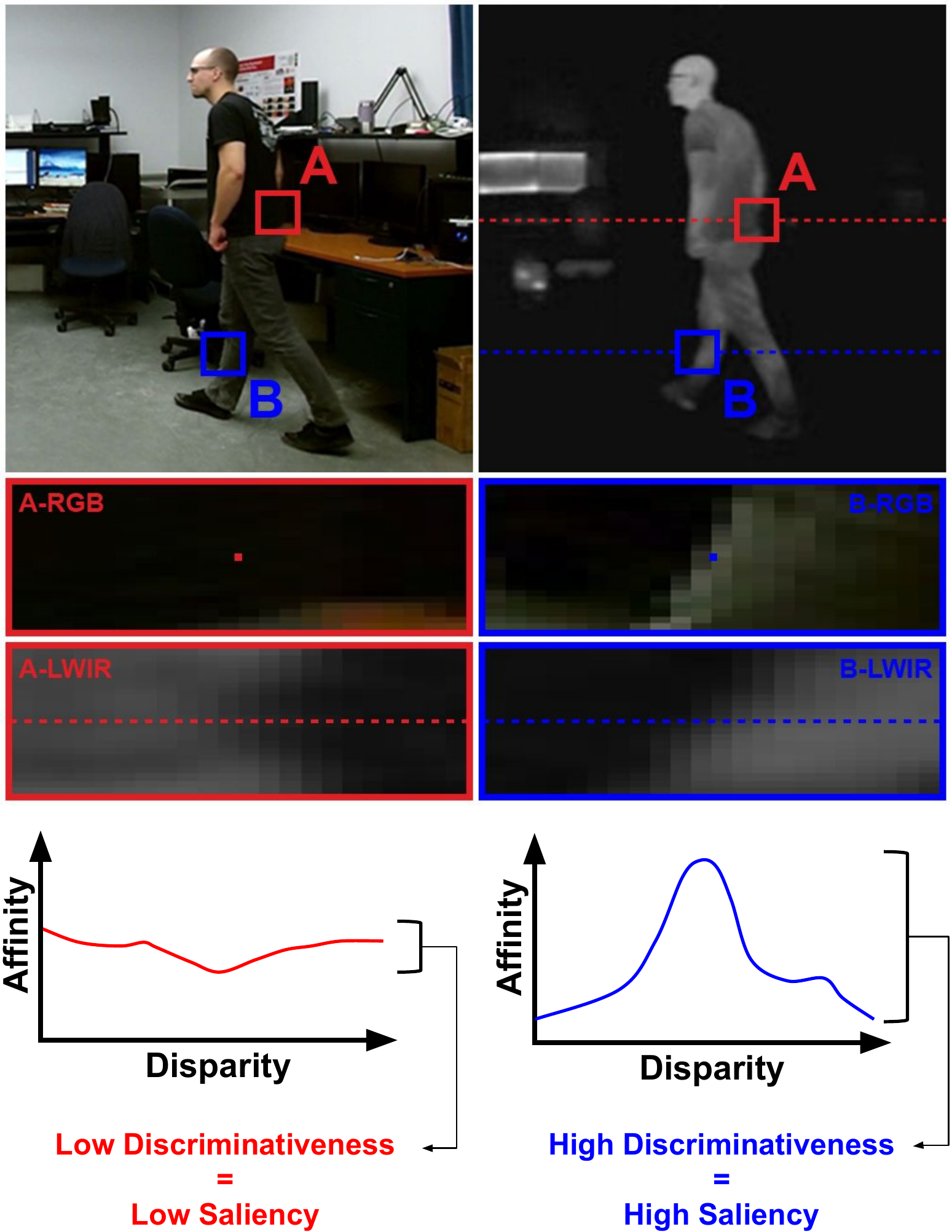}
    \vspace{-3pt}
    \caption{Simplified case of saliency evaluation during a correspondence search on an epipolar line. On the left, for the ``A'' pair, low contrast in one image leads to roughly uniform affinity scores and matching costs, which translate into a low local saliency value. On the right, for the ``B'' pair, good contrast leads to varied affinity scores and matching costs, and a high local saliency value.}
    \label{fig:saliency_epipolar_search_ex}
\end{figure}

\textbf{Appearance and shape terms.} These two terms convey the cost of matching an image patch centered on a pixel $p \in I$ to another one in the second view which is offset according to its disparity label $d_p$. The terms are both defined as
\begin{equation}
    E^{\{\text{appearance, shape}\}}(D) = \sum\limits_{p \in I}\Xagr{A}\big(p,r(p,d_p)\big)\cdot\Xagr{W}\big(p\big),
    \label{eq:stereo_e_data}
\end{equation}
where $r(p,d_p)$ returns the pixel location in the other view obtained by shifting $p$ by $d_p$ on its epipolar line, $\Xagr{A}(p,q)$ encodes the affinity cost for matching descriptor patches centered at $p$ and $q$ in each image, and $\Xagr{W}(p)$ encodes the saliency coefficient for pixel $p$ (detailed further down). For the appearance term, the affinity cost map $\Xagr{A}$ is obtained by densely computing local image descriptors over $I_0$ and $I_1$, and by matching them using L2 distance in 15x15 patches to dampen noise. As stated in Section~\ref{sec:relwork}, classic appearance-based descriptors are not ideal for wide spectrum pairs such as RGB-LWIR. To address this issue, we used Dense Adaptive Self-Correlation descriptors \citep[DASC;][]{kim2015}, which are based on self-similarity measures. We also tested the Local Self-Similarity descriptor \citep[LSS;][]{shechtman2007} during our preliminary experiments, and found a slight decrease in terms of overall registration performance. For the shape term, we densely compute Shape Context descriptors \citep{belongie2002} over $S_0$ and $S_1$, which are the provisional segmentation masks. We then match these descriptors using the same approach as for the appearance term to obtain the shape affinity cost map $\Xagr{A}$. Our hypothesis here is that the combination of these two types of descriptors can provide better matching results than either one alone. However, remember that multispectral matches are often unreliable due to non-discriminative descriptors in uniform image regions or in regions with very low multispectral correlation. To avoid increasing pixel matching penalties in such cases, we multiply the affinity cost by a local saliency coefficient. In both the appearance and shape terms, this local saliency coefficient for a given pixel $p$ is defined as
\begin{equation}
    \Xagr{W}(p) \scalebox{0.85}[1.0]{$\,=\,$}%
    \mmax\!\left\{%
        \Xagr{H}\bigg(\Big[\Xagr{A}\big(p,r(p,d)\big)\,\forall\,d\!\in\!\Xagr{L}_D\!\Big]\bigg),\Xagr{H}\Big(K(p)\Big)\!\right\}\!,%
    \label{eq:stereo_e_data_saliency}
\end{equation}
where $K(p)$ returns the matrix of local descriptors in the patch centered on pixel $p$, and $\Xagr{H}(\cdot)$ computes the sparseness metric of \cite{hoyer2004} over a vector or matrix. This metric returns a value $\in [0,1]$, meaning $\Xagr{W}(p)$ is also in that interval. In simple terms, if all affinity values are uniform (i.e. all disparity offsets have the same cost), and if the local patch's descriptor bins are all uniform, then $\Xagr{W}(p)$ will take a low value. In turn, this will lower the cost for $d_p$ evaluated through the affinity map $\Xagr{A}$, and make local labeling depend more on neighboring decisions through the smoothness term. A simplified case of this is illustrated in Figure~\ref{fig:saliency_epipolar_search_ex}. Besides, note that in $E^{\text{shape}}$, we nullify the saliency outside foreground regions to avoid influencing background disparity estimation around object contours. We can assume that disparity estimation for background regions will be less accurate due to this missing term contribution, but since we focus on the registration of foreground shapes, this is inconsequential. We study the individual contributions of the appearance and shape terms to the overall performance of our approach in Section~\ref{sec:experiments}.

\begin{figure}[t]
    \centering
    \includegraphics[width=0.95\linewidth]{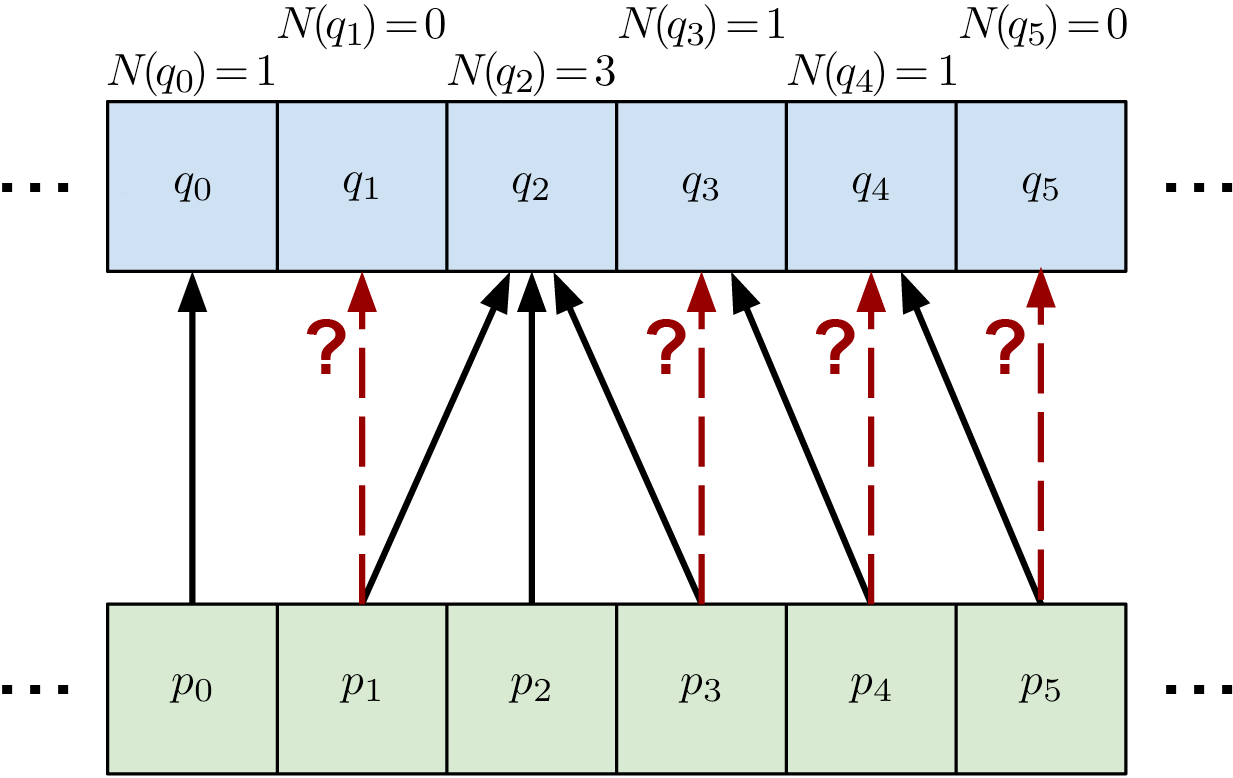}
    \vspace{-3pt}
    \caption{Example showing why an approximation of the uniqueness cost variation must be used under a move-making optimization approach. The two rows are epipolar lines whose pixels have to be matched individually. Already established correspondences are shown with solid black arrows, and move proposals are shown with dashed red arrows. The proposals all originate from a single disparity label for each move, which in this case is $d=0$, meaning $r(p_x,d)=q_x$. Here, the move operation could lower $N(q_2)$ (and thus lower the total energy) by reassociating $p_1$ with $q_1$ and $p_3$ with $q_3$, but the energy variation induced by these swaps cannot be predetermined exactly. It will depend on how many links with $q_2$ are broken during the move (i.e. one or two) due to the other terms, and whether $p_4$ is still linked with $q_4$ afterwards, and so on.}
    \label{fig:uniqueness_ex}
\end{figure}

\textbf{Uniqueness term.} This unary term is used to penalize having multiple epipolar correspondences tied to the same pixel. This helps spread and equalize disparity labels in occluded and weakly discriminative image regions. Our formulation for this term is different from the classic mutual exclusion constraint proposed by \cite{kolmogorov2001}, which assigns an infinite cost to all extra correspondences found for a pixel $p$. Instead, we rely on a soft constraint that permits many-to-one correspondences with gradually increasing costs. This strategy allows our stereo model to temporarily stack extra correspondences during label swaps if the extra cost is worth absorbing. This translates into faster and larger label moves in the early steps of our inference approach, and redistribution of extra correspondence costs over future iterations. Since our method only requires a rough registration of foreground shapes to start properly segmenting them, this allow us to bootstrap the segmentation model without spending too much time on disparity estimation. We define the uniqueness cost for a pixel $p$ as
\begin{equation}
    \Xagr{U}(p) =
        \left\{
            \begin{array}{cl}
                \sum_{n=1}^{N(p)-1} \frac{w{\cdot}n}{w{+}n-1} & \mbox{ if } N(p) > 1 \\
                0 & \mbox{ otherwise}
            \end{array}
        \right.,
    \label{eq:stereo_e_uniqueness_cost}
\end{equation}
where $N(p)$ returns $p$'s current correspondence count with pixels in the other view, and $w$ is a small weight (we used $w{=}3$ in our tests). For this to work, we need to keep track of pixel correspondence counts ($N(p)$) as latent variables in our model. However, since we use a move-making strategy for model inference, many correspondences might be removed in a single iteration. This makes the total cost of a move over several pixels hard to predict with~\eqref{eq:stereo_e_uniqueness_cost} due to its nonlinearity. To solve this problem, we define our uniqueness term as
\begin{equation}
    E^{\text{uniqu.}}(D) = \lambda_u\cdot\sum\limits_{p \in I}\scalebox{1.0}{$\!\left( \frac{\,\text{--}\,\Xagr{U}\big(r(p,d^{\prime}_p)\big)}{N\big(r(p,d^{\prime}_p)\big)}\!+\!\frac{w{\cdot}N\big(r(p,d_p)\big)}{w{+}N\big(r(p,d_p)\big){-}1} \right)\!$},
    \label{eq:stereo_e_uniqueness}
\end{equation}
where $d^{\prime}_p$ is the previous disparity label of $p$, and $\lambda_u$ is a fixed scaling factor. Note that we specify the values used for important factors such as $\lambda_u$ in Section~\ref{sec:approach:inference}, and test their contribution to overall performance in Section~\ref{sec:experiments:params}. The formulation behind~\eqref{eq:stereo_e_uniqueness} provides the worst-case energy variation between two labeling states, and guarantees that estimated label update costs provided to the move-making algorithm will always be similar but greater than the evaluated costs once the full move is complete. The left term of the sum corresponds to the energy refunded if a previous pixel correspondence is broken, and the right term corresponds to an increase due to a new correspondence. The approximation of the true energy variation is required so that the optimizer always minimizes~\eqref{eq:stereo_e_global}, which lets us avoid having to fall back to an older labeling state if the total energy increases. We further explain why this estimation is needed via an example in Figure~\ref{fig:uniqueness_ex}.

\textbf{Smoothness term.} Lastly, we rely on a classic truncated pairwise (first-order) smoothness term to enforce the spatial coherence of our model. This term penalizes cases where neighboring pixels have irregular disparity labels despite being located in a roughly uniform image region, as described by a weak local gradient magnitude. If the gradient detected between the two pixels is instead strong, the penalty is lowered, as object edges more likely correspond with breaks in labeling. We define this term as
\begin{equation}
    E^{\text{\scalebox{0.9}{smooth1}}}(D) = \lambda_{\text{s1}}{\cdot}\!\!\!\!\!\!\sum\limits_{\langle{p,q}\rangle \in \Xagr{N}}\!\!\!\!\min\!\Big(|d_p\smallminus\,d_q|,10\Big)^2\!\!\cdot G^s_{\!I}\!(p,q),
    \label{eq:stereo_e_smooth}
\end{equation}
with
\begin{equation}
    G^s_{\!I}\!(p,q) = \max\!\bigg(\!\exp\!{\Big(1\smallminus\scalebox{1.1}{$\frac{|{\nabla}{I(p{,}q)}|}{g}$}\Big)\!}\,\smallminus\,0.5,\,0\bigg),
    \label{eq:stereo_e_smooth_grad}
\end{equation}
where $\lambda_{\text{s1}}$ is a fixed scaling factor, $\Xagr{N}$ is the set of first order cliques in the graph model, ${\nabla}{I(p{,}q)}$ returns the normalized local image gradient magnitude between pixels $p$ and $q$ of image $I$, and $g$ is a constant value defining the expected object contour gradient magnitude (also specified in Section~\ref{sec:approach:inference}). The truncation value (10 is used) allows large discontinuities to occur by capping the maximum smoothness penalty.

\subsection{Segmentation Model}
\label{sec:approach:segm}

Our segmentation model's role is to integrate multispectral image data so that foreground objects can be properly segmented in both views, even in low contrast imaging conditions. Our model also needs to be lightweight enough so that cost updates and inference is fast, as shape priors are continuously modified. Since our goal is to build an unsupervised approach, we initialize the priors described below using the approximate masks provided by a monocular segmentation method \citep[i.e. the one of][]{pawcs2016}. This method was chosen because it can detect multiple foreground objects at once, and it can keep segmenting them at least partially if they become immobile. In Section~\ref{sec:experiments}, we show that our method works even when an initialization mask is provided for only one of the two views.

\begin{figure}[t]
    \centering
    \includegraphics[width=\linewidth]{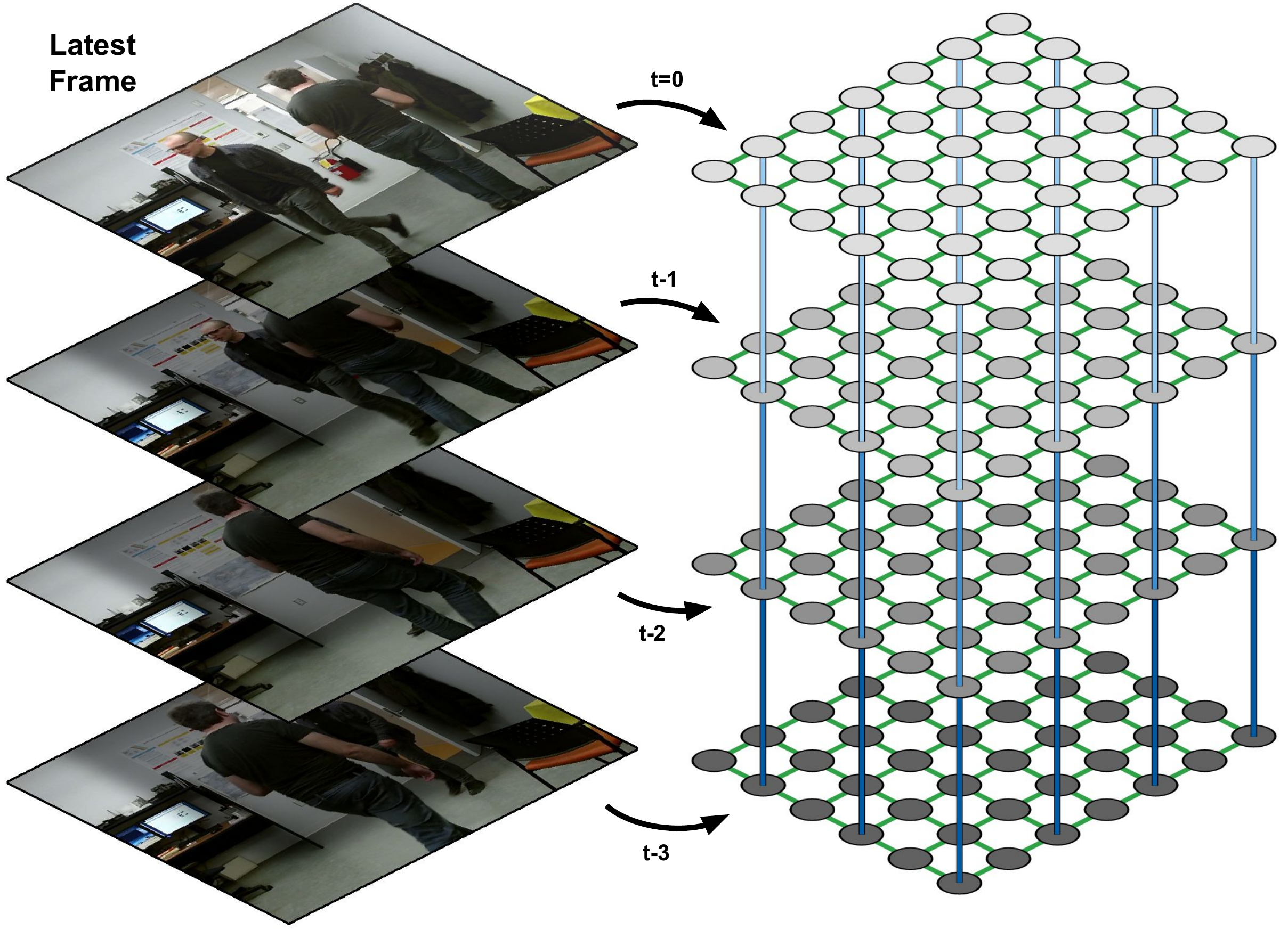}
    \vspace{-12pt}
    \caption{Illustration of the simplified frame layering used in our segmentation model for temporal labeling refinement. In green, the first-order cliques that form $E^{\text{smooth2}}$ are used to enforce spatial coherence in every layer. In blue, the higher order cliques that form $E^{\text{temp}}$ are used to enforce temporal coherence across layers. Note that due to foreground motion, these cliques would not all be linked to the same underlying nodes; in reality, the links are dictated by image realignment based on optical flow.}
    \label{fig:temporal_layering_ex}
\end{figure}

We describe the energy cost of a pixel-wise segmentation proposal $S$ as
\begin{equation}
    \begin{aligned}
        E^{\text{segm}}(S) ={} & E^{\text{color}}(S)\,+ E^{\text{contour}}(S) \\
                               & \,+ E^{\text{smooth2}}(S)\,+ E^{\text{temp}}(S)
    \end{aligned}
    \label{eq:segm_e_global}.
\end{equation}
Once again, the terms of this cost function are defined so that various characteristics expected of the segmentation masks can be promoted. The first two terms are unary costs summed over all pixels, and their role is to influence local segmentation decisions based on image data. The color data term maximizes the separation between the color distributions of foreground and background pixels, while the contour data term penalizes shape mismatches between the views based on distance transforms. The third term is a pairwise smoothness sum similar to~\eqref{eq:stereo_e_smooth}, and is used to penalize labeling irregularities in uniform image regions. Lastly, the temporal term is a sum of higher order clique costs used to enforce temporal labeling coherence. These terms are all described in the following paragraphs. Note that due to the presence of the higher order temporal term in~\eqref{eq:segm_e_global}, our model is built as a multi-layer lattice, as illustrated in Figure~\ref{fig:temporal_layering_ex}. The top layer's nodes correspond to the pixels of the latest frame of the video sequence, and lower layers' nodes correspond to the pixels of older frames. This effectively creates a pipeline where segmentation masks can be improved over time based on new image data. We discuss the improvement achieved using this approach with various pipeline depths in Section~\ref{sec:experiments}.

\textbf{Color term.} We define the cost for this unary term using a color mixture model for each modality of the stereo pair. We employ the classic approach of \cite{rother2004} which relies on Gaussian mixture models to represent foreground and background regions. These models can provide us with the probability that a pixel belongs to the background or foreground based on its color value. In our implementation, we use six mixture components, and use our initial and updated segmentation masks to refine our models after each iteration, in each frame. We define the color cost of all pixels as
\begin{equation}
    E^{\text{color}}(S)\! = \!\sum\limits_{p \in I}
        \left\{
            \begin{array}{cl}
                \!\!\!{-}\log\Big(h(I_p;\bm{\beta_1}{,}\bm{\mu_1}{,}\bm{\Sigma_1})\Big) & \mbox{ if } s_p = 1 \\
                \!\!\!{-}\log\Big(h(I_p;\bm{\beta_0}{,}\bm{\mu_0}{,}\bm{\Sigma_0})\Big) & \mbox{ otherwise}
            \end{array}
        \right.\!\!,
    \label{eq:segm_e_color}
\end{equation}
where $h(x;\bm{\beta},\bm{\mu},\bm{\Sigma})$ returns the relative likelihood that the pixel color $x$ fits a Gaussian mixture model with component weights $\bm{\beta}$, means $\bm{\mu}$ and covariance matrices $\bm{\Sigma}$. Note that the  parameter subscripts in~\eqref{eq:segm_e_color} indicate that either the foreground or background model is used based on $s_p$. These parameters are initialized using \emph{k-means}, and refitted after every minimization step using the new estimated segmentation masks.

\textbf{Contour term.} Next, we define another data term that penalizes label swaps far from shape boundaries, and that combines these boundaries across the stereo pair. Its value is computed using shape distance transforms: first, we build maps in which each pixel is assigned its Euclidean distance to the closest pre-existing foreground or background pixel in the current view. We then use these maps to deduce the label update costs for each pixel in our graph, considering a mix of distances in both views at once (note the use of subscript $k$ below). More specifically, we define our contour term as
\begin{equation}
    \!E_k^{\text{cont.}}(S_k)\!=\!\lambda_c\!\!\cdot\!\!\sum\limits_{p \in I_k}\!
        \left\{\!
            \begin{array}{cl}
                \scalebox{0.97}{$\!F_k\!\big(p\big){+}\lambda_m{\cdot}F_{k'}\!\big(r(p,d_p)\big)$} & \mbox{\scalebox{0.93}{if}}\,\,\,s_{p}\,{=}\,1 \\
                \scalebox{0.97}{$\!B_k\!\big(p\big){+}\lambda_m{\cdot}B_{k'}\!\big(r(p,d_p)\big)$} & \mbox{\scalebox{0.93}{otherwise}}
            \end{array}
        \right.\!\!\!\!,
    \label{eq:segm_e_contour}
\end{equation}
where $\lambda_c$ and $\lambda_m$ are fixed scaling factors, $k'$ is the opposite index of $k$ in the stereo pair, $F_k(p)$ returns a nonlinear distance cost (described below) for pixel $p$ based on its distance to the closest foreground pixel in the previous segmentation of view $k$, and similarly for $B_k(p)$ with background pixels. Note in~\eqref{eq:segm_e_contour} that $\lambda_m$ scales the term's multispectral cost contribution. During our tests, we give it a value $\in \interval[open]{0}{1}$, meaning that shape contours will prefer sticking to their own previous results. This improves the stability of the segmentation while optimizing, reducing the risk of eliminating relevant shape fragments too rapidly. For the nonlinear distance cost function behind $F_k(p)$ and $B_k(p)$, we use an exponential to increase the contrast between close range and long range contour overlaps. More specifically, we use a relation of the form
\begin{equation}
    \text{distance-cost}(p) \propto \frac{1}{\exp\!\big(\!-t(p)\big)}
    \label{eq:segm_e_contour_dist},
\end{equation}
where $t(p)$ returns the actual Euclidean distance between $p$ and its nearest pixel with a foreground or background label in the previous inference result, depending on the current value of $s_{p}$. The contour term's main responsibility is to control the evolution of object contours over several optimization passes. The multispectral contribution allows contours to be modified in regions where only one modality contributes meaningful information. The simple formulation of our contour term also avoids the needless filling of cavities, and it makes no assumption on the foreground-to-background ratio in the images.

\textbf{Smoothness term.} This pairwise term is similar to the one used in~\eqref{eq:stereo_e_smooth}; its role is to penalize label discontinuities everywhere except for image regions where local gradients are strong. In this case, however, we reuse the multispectral contribution idea of~\eqref{eq:segm_e_contour}, and apply it to the gradient scaling factor. We define this term as
\begin{equation}
    \!E_k^{\text{\scalebox{0.85}{smooth2}}}(S_k)\,\smallequal\,\lambda_{\text{s2}}\,{\cdot}\!\!\!\!\!\!\!\!\!\sum\limits_{\scalebox{0.7}{$\langle{p,q}\rangle \in \Xagr{N}_k$}}\!\!\!\!\!\!\!\scalebox{0.9}{$\Big(\!s_{p}\scalebox{0.9}{$\xor$}s_{q}\!\Big){\cdot}\Big(\!G^s_{\!I_{k}}\!\big(p{,}q\big)\,\smallplus\,{\lambda_m}\!{\cdot}G^s_{\!I_{k'}}\!\big(p'{,}q'\big)\!\Big)$},
    \label{eq:segm_e_smooth}
\end{equation}
where $\lambda_{\text{s2}}$ is a fixed scaling factor, $\xor$ is the XOR operator, $p'$ is a shorthand for $r(p,d_p)$, and $q'$ is a shorthand for $r(q,d_q)$. In~\eqref{eq:segm_e_smooth}, the right-hand parentheses group returns the gradient coefficient with its multispectral contribution, and the left-hand group returns 1 or 0 based on whether a label discontinuity is found. As before, the use of local image gradients helps ``snap'' these discontinuities to real object contours. However, the multispectral contribution allows shape contours to settle in uniform regions if the other view possesses a strong local gradient there. Paired with the contour term, this allows our model to properly expand and contract shape boundaries across image modalities. We study the effect of $\lambda_m$ on the performance of our method in Section~\ref{sec:experiments}.

\textbf{Temporal term.} Lastly, we present the formulation and role of our temporal term. Unlike the other terms presented so far, this term is based on higher order cliques that are composed at runtime, and updated for each frame. The role of these cliques is to enforce spatiotemporal labeling coherence despite foreground object motion. Our graph structure can be visualized as a stack of analyzed frames; this structure is shown in Figure~\ref{fig:temporal_layering_ex}. While the depth (or layer count) of this stack is predetermined, its temporal cliques are composed based on node realignments provided by optical flow maps. This allows cliques to remain attached to the same object part despite movement, and thus enforce labeling smoothness across frames. We compute optical flow maps using the method proposed by \cite{kroeger2016}. As for the cost term itself: given $\Xagr{C}$, the set of all temporal cliques in our model, and using the subscript $l$ to identify different temporal layers in these cliques, we define it as
\begin{equation}
    E^{\text{temp}}(D) = \lambda_{\text{s2}}{\cdot}\!\!\sum\limits_{c\,\in\,\Xagr{C}}\sum\limits_{l=1}^{L-1}\!\big(s_{c{,}l}\,\scalebox{0.9}{$\xor$}\,s_{c{,}(l+1)}\big) \cdot G^t\!\big(c,l\big),
    \label{eq:segm_e_temporal}
\end{equation}
where $\lambda_{\text{s2}}$ is the same scaling factor as in~\eqref{eq:segm_e_smooth}, $L$ is the pipeline's depth in frames, $s_{c,l}$ returns the label of the $l$\nobreakdash-th node in clique $c$, and $G^t\!(c,l)$ returns a scaling factor for clique $c$ at layer $l$ (described next). Overall, this term is similar to the pairwise smoothness terms described earlier, but it can link more nodes together. However, instead of scaling costs via local image gradients, we rely on temporal image gradients in $G^t\!(c,l)$. These new gradient values are obtained by computing the absolute color differences between pixels of consecutive frames realigned using optical flow maps. Strong color differences are indicative of uncertain regions where consistency costs should be reduced due to occlusions or bad optical flow estimation. Here, similarly to~\eqref{eq:stereo_e_smooth_grad}, we define the new gradient scale term as
\begin{equation}
    G^t(c,l) = \max\!\bigg(\!\exp\!{\Big(1\smallminus\scalebox{1.2}{$\frac{|i_{c{,}l} - i_{c{,}(l{+}1)}|}{g}$}\Big)\!}\,\smallminus\,0.5,\,0\bigg),
    \label{eq:segm_e_temporal_grad}
\end{equation}
where $i_{c,l}$ returns the color value of the $l$\nobreakdash-th node in clique $c$. We study the contribution of this new term in Section~\ref{sec:experiments} given various layer count configurations.

\subsection{Inference and Implementation Details}
\label{sec:approach:inference}

Simultaneously minimizing the cost functions defined in~\eqref{eq:stereo_e_global} and~\eqref{eq:segm_e_global} is not trivial. Both functions rely on each other's provisional results as dynamic priors, and~\eqref{eq:segm_e_global} contains a higher order term. A simple cost function can typically be minimized iteratively using a move-making algorithm such as $\alpha$\nobreakdash-expansion \citep{boykov2001} that returns a local minimum within a known factor of the global minimum. In our case, the dynamic weights and links used to connect our two cost functions cause their global objective to be updated each time a new labeling is obtained for either half of the model. This means that the global minimum of our model is always changing, and that reaching it is difficult. Instead, we focus on converging to a local minimum in each function by alternating label move operations. Recently proposed move-making algorithms can deal with higher-order terms and dynamic priors without having to resort to a move-and-check or rollback strategy \citep[c.f.][]{lempitsky2010,kappes2013}. However, to reach a local minimum in both functions simultaneously, the terms have to be carefully designed so that the cost functions can converge under roughly similar conditions. We achieve this as anticipated using shape contours: these tend to settle on the maxima in gradient intensity maps that correspond to object boundaries, and can be easily matched across image modalities despite some local shape variability. In practice, our optimization strategy converges once the target objects in the scene (roughly identified via the initialization masks) are properly covered by foreground segments that are registered between the two views. This convergence also happens without having to use a decaying metaparameter to force a solution after a fixed number of iterations.

We rely on the move-making algorithms of \cite{fix2011} and \cite{fix2014} for the inference of our stereo and segmentation models, respectively. Both are modified for use in a dynamic graph structure. While faster inference solutions do exist, these were deemed fast enough for our experiments, even without having to parallelize label moves. In both cases, our move proposals only consist of uniform labeling maps, meaning our inference approach is fairly similar to $\alpha$\nobreakdash-expansion. We build our graphical models in C++ using the OpenGM library \citep{opengm}, and reuse the same structure for all frames in a video, updating only the composition of temporal factors in~\eqref{eq:segm_e_temporal} as required. We settled for these two generic optimizers to show that the formulation of our models is not tied to the optimization approach we use.

We tackle the alternating minimization of energies \eqref{eq:stereo_e_global} and~\eqref{eq:segm_e_global} for each frame of a video by first minimizing the stereo model's energy using unary terms only, or by realigning its previous disparity labeling result via optical flow. Simultaneously, the segmentation model is initialized using the masks provided by an unsupervised monocular method, as stated earlier. Then, segmentation and disparity label moves are iteratively computed in small batches until no more moves in $\Xagr{L}_S$ can reduce the energy of~\eqref{eq:segm_e_global}. This typically happens after less than three passes over the disparity label space ($\Xagr{L}_D$), and less than 50 moves in the segmentation label space ($\Xagr{L}_S$), the exact number depending on the quality of the initialization. For reference, with our baseline implementation, this is equivalent to approximately 30 seconds worth of processing time on a single core of a 3.7~GHz Intel i7-8700K processor for a VGA-sized image pair. This processing time seems to scale in a roughly linear fashion with respect to the number of pixels in the analyzed images.

As for the free parameters listed earlier, we use the following configuration for our tests in the next section:
\begin{itemize}
  \item Stereo model uniqueness term weight: $\lambda_u = 0.4$
  \item Stereo model smoothness term weight: $\lambda_{\text{s1}} = 0.001$
  \item Expected object contour gradient intensity: $g = 30$
  \item Segmentation model contour term weight: $\lambda_c = 7$
  \item Segmentation model smoothness weight: $\lambda_{\text{s2}} = 7$
  \item Multispectral contribution term weight: $\lambda_m = 0.5$
\end{itemize}
The values listed above have been empirically found to provide good overall segmentation performance on a small subset of our test data via grid search. As previously noted, we study the effect of several of these parameters on the overall performance of our method in the next section. For optical flow and DASC descriptors computations, we kept the default parameters provided by their original authors. For Shape Context computations, we used 50 pixel-wide descriptors with 10 angular bins and 3 radial bins. For the depth of our frame processing pipeline, we used two temporal layers (i.e. the current frame and the previous one), as adding more did not improve overall performance significantly over the extra processing cost; this is discussed in Section~\ref{sec:experiments:params}. Finally, to reduce the computational cost when using higher order terms in our segmentation model, we use a stride of two pixels when creating the temporal cliques used in~\eqref{eq:segm_e_temporal}. For more implementation details, we refer the reader to our source code\footnote{\url{https://github.com/plstcharles/litiv}}.

\section{Experiments}
\label{sec:experiments}

\begin{table*}[!t]
    \centering
    \rowcolors{3}{gray!15}{white}
    \begin{threeparttable}
        \renewcommand{\arraystretch}{1.05}
        \caption{Evaluation results on the multispectral video segmentation dataset of \cite{palmero2016}. Bold results are the best in that category across all methods.}
        \begin{tabular}{| M{3.2cm} | M{1.0cm} || M{1.0cm} M{1.0cm} | M{1.0cm} M{1.0cm} | M{1.0cm} M{1.0cm}  M{1.2cm} |}
            \hline
            \multirow{2}{*}[-4pt]{\bf{Method}} & \multirow{2}{*}[-4pt]{\bf{Metric}} & \multicolumn{2}{c|}{\multirow{2}{*}[1pt]{\bf{Scene 1}}} & \multicolumn{2}{c|}{\multirow{2}{*}[1pt]{\bf{Scene 3}}} & \multicolumn{3}{c|}{\multirow{2}{*}[1pt]{\bf{Overall}}} \\[5pt]
                     &           &  visible & LWIR & visible & LWIR & visible & LWIR & \bf{Average} \\
            \hline
            \cellcolor{white}    & \cellcolor{white}\emph{Pr} & 0.820 & 0.755 & 0.716 & 0.514 & 0.768 & 0.635 & 0.701 \\
            \cellcolor{white}    & \cellcolor{white}\emph{Re} & 0.810 & \bf{0.975} & 0.688 & \bf{0.969} & 0.749 & \bf{0.972} & 0.861 \\
            \cellcolor{white}\multirow{-3}{*}{\specialcell{\cellcolor{white}\cite{pawcs2016}\\\cellcolor{white}(unsupervised)}}
                                 & \cellcolor{white}$F_1$     & 0.815 & 0.851 & 0.702 & 0.672 & 0.758 & 0.762 & 0.760 \\
            \hline
            \cellcolor{white}    & \cellcolor{white}\emph{Pr} & --- & --- & \bf{0.817} & 0.777 & --- & --- & --- \\
            \cellcolor{white}    & \cellcolor{white}\emph{Re} & --- & --- & 0.568 & 0.564 & --- & --- & --- \\
            \cellcolor{white}\multirow{-3}{*}{\specialcell{\cellcolor{white}\cite{palmero2016}\\\cellcolor{white}(semi-supervised)}}
                                 & \cellcolor{white}$F_1$     & --- & --- & 0.670 & 0.654 & --- & --- & --- \\
            \hline
            \cellcolor{white}    & \cellcolor{white}\emph{Pr} & 0.685 & 0.808 & 0.653 & \bf{0.847} & 0.669 & \bf{0.828} & 0.748 \\
            \cellcolor{white}    & \cellcolor{white}\emph{Re} & 0.759 & 0.896 & \bf{0.929} & 0.916 & 0.844 & 0.906 & 0.875 \\
            \cellcolor{white}\multirow{-3}{*}{\specialcell{\cellcolor{white}\cite{rother2004}\\\cellcolor{white}(GrabCut; supervised)}}
                                 & \cellcolor{white}$F_1$     & 0.721 & 0.850 & 0.767 & \bf{0.880} & 0.744 & \bf{0.865} & 0.804 \\
            \hline
            \cellcolor{white}    & \cellcolor{white}\emph{Pr} & \bf{0.894} & \bf{0.860} & 0.788 & 0.749 & \bf{0.841} & 0.804 & \bf{0.821} \\ 
            \cellcolor{white}    & \cellcolor{white}\emph{Re} & \bf{0.902} & 0.901 & 0.918 & 0.937 & \bf{0.910} & 0.919 & \bf{0.914} \\ 
            \cellcolor{white}\multirow{-3}{*}{\specialcell{\cellcolor{white}Proposed method\\\cellcolor{white}(unsupervised)}}
                                 & \cellcolor{white}$F_1$     & \bf{0.898} & \bf{0.880} & \bf{0.848} & 0.833 & \bf{0.873} & 0.857 & \bf{0.866} \\ 
            \hline
        \end{tabular}
        \label{tab:segm:comp}
    \end{threeparttable}
\end{table*}

In this section, we first discuss our evaluation methodology, and then present evaluation results for mutual segmentation and stereo registration. Since close-range (non-planar) multispectral video datasets are quite uncommon in the literature, we had to adapt existing datasets to our problem. For multispectral mutual segmentation, we rely on a modified version of the VAP trimodal dataset of \cite{palmero2016}; the modifications we made are detailed in Section~\ref{sec:experiments:vap}. For stereo registration, we rely on the benchmark of \cite{bilodeau2014}. We follow up with an ablation study of our method in which we remove key terms from our energy functions, and then study the effect of tuning key parameters of these terms. Finally, we provide evaluation results for both segmentation and stereo registration on a newly captured and annotated RGB-LWIR dataset for future comparisons.

\subsection{Evaluation Methodology}
\label{sec:experiments:evalmethod}

Since our primary goal is mutual foreground segmentation, we employ binary classification metrics for the first part of our evaluation. Commonly used metrics in the context of video segmentation are Precision (\emph{Pr}), Recall (\emph{Re}), and $F_1$ score. These are based on three types of pixel-wise classification result counts, namely True Positives (TP), False Positives (FP), and False Negatives (FN). These metrics are defined as
\begin{equation}
    \text{Precision} = \frac{\text{TP}}{\text{TP}+\text{FP}},
    \label{eq:eval_pr}
\end{equation}
\begin{equation}
    \text{Recall} = \frac{\text{TP}}{\text{TP}+\text{FN}},
    \label{eq:eval_re}
\end{equation}
\begin{equation}
    F_1 = \frac{2{\cdot}\text{Precision}{\cdot}\text{Recall}}{\text{Precision}+\text{Recall}},
    \label{eq:eval_fm}
\end{equation}
In all three cases, higher values indicate better performance. The $F_1$ score corresponds to the harmonic mean of the precision and recall scores. We use it as an overall indicator of binary segmentation performance, as it was shown in the work of \cite{cdnet2012} to be strongly correlated with the final ranking of methods on a large binary segmentation dataset based on numerous other metrics.

Our second goal is to evaluate stereo registration performance. For this, we employ the strategy of the Middlebury dataset \citep{scharstein2014}, and report the percentage of pixels labeled with disparity errors larger than some fixed distance thresholds (in pixels). We also report average frame-wide pixel disparity errors, noted $\bar{d}_{\text{err}}$ below. In this case, lower values indicate better performance.

\subsection{VAP 2016 Dataset}
\label{sec:experiments:vap}

\begin{figure*}[t]
    \centering
    \hspace{-2mm}
    \begin{tabular}{c@{\hskip 1mm}c@{\hskip 1mm}c@{\hskip 1mm}c}
        \figResultsRowWideHeader{Segmentation results of \cite{pawcs2016}}{Segmentation results of the proposed method}
        \figResultsRowHeader{RGB}{LWIR}{RGB}{LWIR}
        \figResultsRow{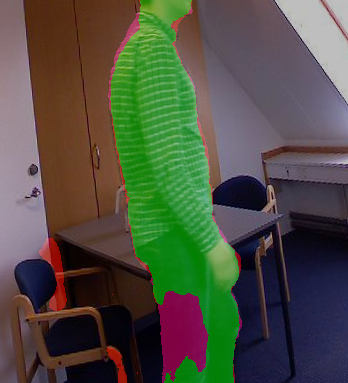}{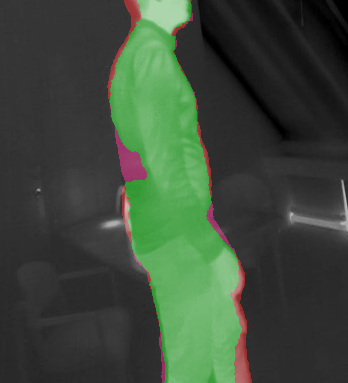}{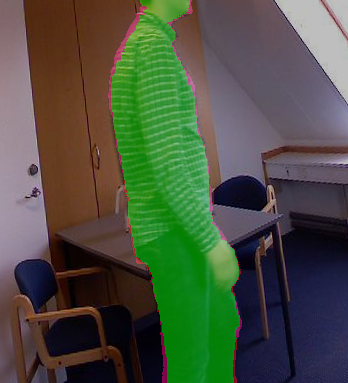}{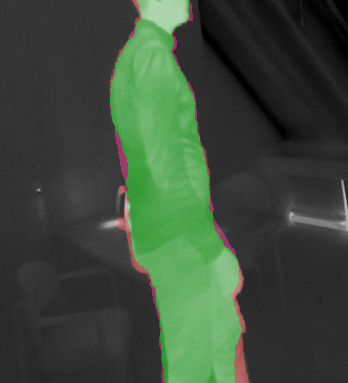}
        \figResultsRow{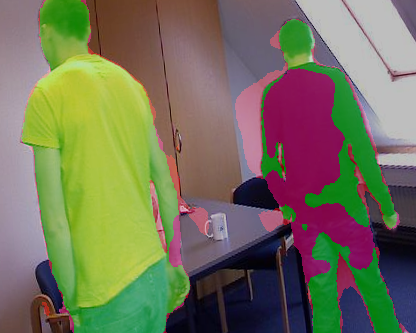}{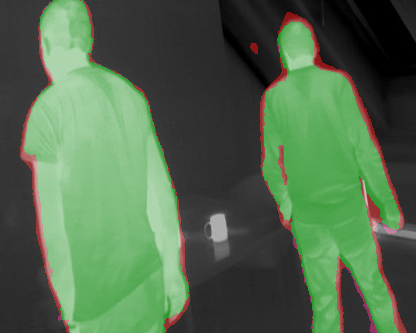}{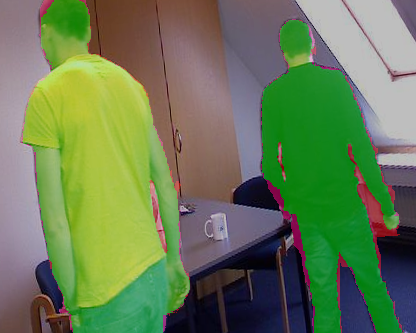}{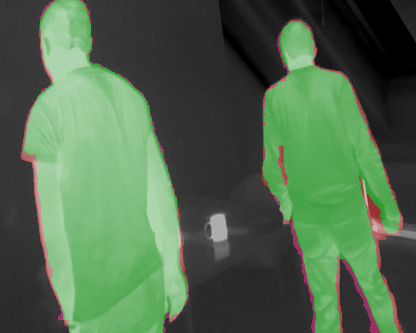}
        \figResultsRow{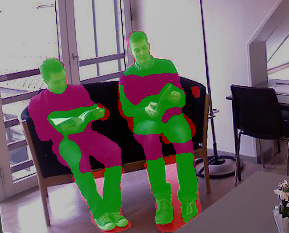}{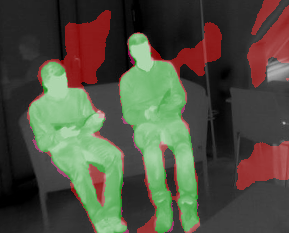}{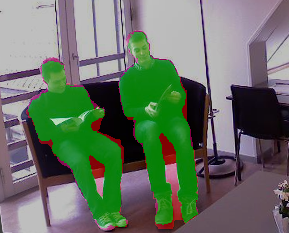}{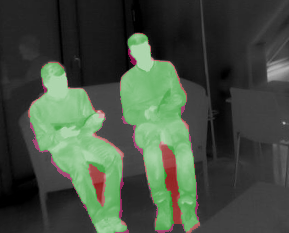}
        \figResultsRow{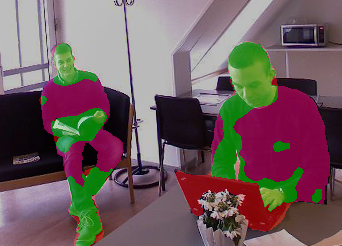}{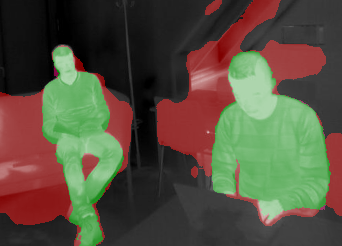}{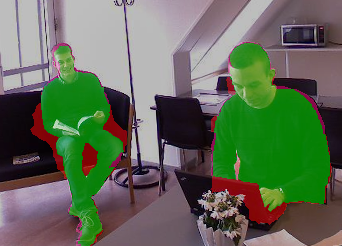}{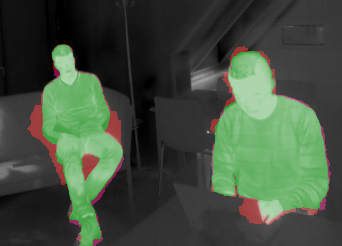}
    \end{tabular}
    \vspace{-3pt}
    \caption{Examples of typical segmentation results from the VAP dataset of~\cite{palmero2016}; the left two columns show the segmentation masks obtained via the method of \cite{pawcs2016} and used to initialize our method, and the right two columns show our final segmentation masks. Image regions properly classified as foreground are highlighted in green over the original images, while regions highlighted in orange and magenta show false positives and false negatives, respectively. Images have been cropped to show more details.}
    \label{fig:res:vap2016}
\end{figure*}

For this first part of our evaluation, we adapted the dataset of \cite{palmero2016} to our needs. This dataset was originally intended for the trimodal (RGBD-LWIR) detection and segmentation of people in images, and it is provided as a set of videos. It consists of 5724 image triplets split into three scenes, with their associated groundtruth foreground-background segmentation masks. We obtained the calibration data used by the original authors to roughly register scene contents via homographies, and rectified all RGB and LWIR image pairs using the OpenCV calibration toolbox. The depth images were left unused during all our experiments, and the second scene was removed due to missing calibration data. Finally, to avoid skewing the performance evaluation by continuously segmenting empty frames or frames with purely static and/or unoccluded foreground regions, we manually selected a subset of groundtruth masks for our experiments. These masks were picked at a rate of roughly 2 Hz from all originally available masks while focusing on time spans with people interacting.

We present the segmentation performance of our proposed method, as well as the performance of baseline video and image segmentation methods in Table~\ref{tab:segm:comp}. We could not evaluate the performance of the works listed in the last paragraph of Section~\ref{sec:relwork} that simultaneously tackle segmentation and registration due to a lack of open-source code and datasets. Besides, comparing our results to those of other methods that assume single-spectrum data or planar scenes would also be unfair. For the video segmentation baseline, we rely on the method of \cite{pawcs2016}, which is fully unsupervised. We use the method's default parameters from its original implementation, and process each spectrum individually. For the image segmentation baseline, we rely on the GrabCut method of \cite{rother2004}, and provide this method with manually defined bounding boxes for all foreground objects. We used OpenCV's GrabCut implementation, and ran five iterations per image. Finally, we provide partial results for the method of~\cite{palmero2016} that were obtained using the original predictions provided by the authors.

\begin{table*}[t]
    \centering
    \rowcolors{3}{gray!15}{white}
    \begin{threeparttable}
        \renewcommand{\arraystretch}{1.2}
        \caption{Evaluation results on the multispectral video registration dataset of~\cite{bilodeau2014}. Bold results are the best in that category across all methods.}
        \begin{tabular}{| M{3.3cm} || M{0.7cm} M{0.7cm} M{0.7cm} |  M{0.7cm} M{0.7cm} M{0.7cm} | M{0.7cm} M{0.7cm} M{0.7cm} | M{0.7cm} M{0.7cm} M{0.7cm} |}
            \hline
            \multirow{2}{*}[-8pt]{\bf{Method}} & \multicolumn{3}{c|}{\multirow{2}{*}[1pt]{\bf{Video 1}}} & \multicolumn{3}{c|}{\multirow{2}{*}[1pt]{\bf{Video 2}}} & \multicolumn{3}{c|}{\multirow{2}{*}[1pt]{\bf{Video 3}}} & \multicolumn{3}{c|}{\multirow{2}{*}[1pt]{\bf{Overall}}} \\[5pt]
                                 & \specialcell{\% err.\\[-4pt]\scalebox{0.9}{\textgreater 1px}} & \specialcell{\% err.\\[-4pt]\scalebox{0.9}{\textgreater 4px}} & $\bar{d}_{\text{err}}$ & \specialcell{\% err.\\[-4pt]\scalebox{0.9}{\textgreater 1px}} & \specialcell{\% err.\\[-4pt]\scalebox{0.9}{\textgreater 4px}} & $\bar{d}_{\text{err}}$ & \specialcell{\% err.\\[-4pt]\scalebox{0.9}{\textgreater 1px}} & \specialcell{\% err.\\[-4pt]\scalebox{0.9}{\textgreater 4px}} & $\bar{d}_{\text{err}}$ & \specialcell{\% err.\\[-4pt]\scalebox{0.9}{\textgreater 1px}} & \specialcell{\% err.\\[-4pt]\scalebox{0.9}{\textgreater 4px}} & $\bar{d}_{\text{err}}$ \\
            \hline
            \cellcolor{white}\scalebox{0.9}{OpenCV's Block Matcher}
                & 95.5 & 95.3 &27.51 & 99.8 & 99.7 &38.99 & 99.8 & 99.6 &34.76 & 98.3 & 98.2 &33.75 \\
            \cellcolor{white}LSS Sliding Window
                & 69.1 & 45.7 & 8.50 & 89.4 & 77.3 & 9.87 & 73.6 & 36.0 & 7.50 & 77.4 & 53.0 & 8.62 \\
            \cellcolor{white}MI Sliding Window
                & 82.0 & 62.5 &10.20 & 86.9 & 61.4 & 9.78 & 84.4 & 63.2 &10.08 & 84.4 & 62.4 &10.02 \\
            \cellcolor{white}DASC Sliding Window
                & 79.2 & 55.0 & 8.94 & 77.4 & 55.9 & 9.11 & 73.5 & 42.1 & 6.88 & 76.7 & 51.0 & 8.31 \\            
            \cellcolor{white}Proposed method
                & \bf{47.7} & \bf{17.3} & \bf{3.28} & \bf{55.6} & \bf{25.4} & \bf{3.11} & \bf{52.0} & \bf{17.5} & \bf{3.26} & \bf{51.8} & \bf{20.1} & \bf{3.21} \\ 
            \hline
        \end{tabular}
        \label{tab:registr:comp}
    \end{threeparttable}
\end{table*}

We can observe that our proposed method outperforms the unsupervised video segmentation approach of~\cite{pawcs2016} in terms of overall $F_1$ score by a margin of 0.1, equal to a relative improvement of over 13\%. This confirms that our approach can properly integrate multispectral information through stereo registration in order to improve segmentation performance beyond that of a state-of-the-art monocular method. Interestingly, our proposed method even outperforms the supervised image segmentation approach of \cite{rother2004}, which relies on manual annotations to pinpoint all foreground objects in every frame. This can be explained by the fact that foreground objects in this dataset have better contrast in the LWIR spectrum than in the visible spectrum, and because our approach propagates this contrast information across the stereo pair. Additionally, our method outperforms the semi-supervised approach of~\cite{palmero2016} in Scene 3 despite having to estimate full disparity maps for stereo registration, and without requiring training. Finally, we show in Figure~\ref{fig:res:vap2016} some qualitative results for this dataset. The last row of this figure presents an interesting case: in this frame pair, the initial foreground masks provided to our method both contain important errors in different regions, but the output is excellent. This shows that despite not having a proper foreground shape template, the real underlying shape can be found and extracted correctly via our iterative process.

\subsection{Bilodeau~\emph{et al.} 2014 Dataset}
\label{sec:experiments:bilodeau2014}

We now evaluate our proposed method's stereo registration accuracy using the benchmark dataset of \cite{bilodeau2014}. This dataset was originally intended for the evaluation of image descriptors and similarity measures in the context of multispectral stereo matching, once again provided as a set of videos. It consists of 5390 RGB-LWIR frame pairs split into three scenes, with over 25,000 sparse correspondences annotated on visible foreground objects.

As stated before, we evaluate performance on this dataset by analyzing the accuracy of disparity labelings. Unfortunately, previous works tackling multispectral registration have often relied on their own foreground overlap ratios to assess their performance \citep[e.g.][]{nguyen2016}, meaning comparisons here are impossible. Here, to provide a reusable evaluation baseline, we compare our results to those obtained using a sliding window patch-matching approach, similar to the strategy used by \cite{bilodeau2014}. In short, local disparity labels are assigned based on the best match (or smallest distance) found between image patches in a winner-takes-all fashion. To describe the similarity between these image patches, we rely on descriptors, namely LSS \citep{shechtman2007} and DASC \citep{kim2015}, and on Mutual Information scores \citep[MI;][]{maes1997}. Note that for these experiments, we used the same metaparameters (e.g. patch size, bin counts) as those used by our own method, or translated them to be roughly equivalent. Also, for fairness, we relied on the same smoothness term we used in our own method ($E^{\text{\scalebox{0.9}{smooth1}}}$) to regularize the patch matching disparity estimation results. Finally, to highlight the issue of applying traditional stereo registration methods on multispectral datasets, we evaluate the block matching algorithm of K. Konolige implemented in OpenCV. These results are presented in Table~\ref{tab:registr:comp}.

We can note that our proposed method performs very well compared to the baseline methods. Unsurprisingly, OpenCV's block matching method fails on this dataset as it tries to compare image textures directly across the pair, despite their low correlation. The approaches based on self-similarity descriptors (LSS, DASC) and mutual information perform slightly better, but still produce highly inaccurate results. On average, above 50\% of all the evaluated points are labeled with disparities at least four pixels off from the groundtruth. On the other hand, our approach manages to label 51.8\% of all evaluated points within a single pixel of the groundtruth, and provides an average disparity error of only 3.21 pixels. Note however that while this performance is good enough for our primary task (mutual foreground segmentation), it is still far from the current state-of-the-art in single-spectrum stereo registration. For example, on the Middlebury dataset \citep{scharstein2014}, top-performing methods typically label less than 20\% of all points with a disparity error larger than a single pixel. This highlights the difficulty of multispectral stereo registration.

\subsection{Parameters and Ablation Study}
\label{sec:experiments:params}

In this section, we study the behavior of our method when key terms and parameters are modified from the default configuration listed in Section~\ref{sec:approach:inference} on the two previously introduced datasets. First, we perform an ablation study to determine which energy terms are the most important in our models; this study is presented in Table~\ref{tab:ablation_study}.

According to the $F_1$ scores, modifying the stereo energy formulation only has a small effect on segmentation performance. On the other hand, removing the color or contour terms from the segmentation energy has larger impacts, and the latter of the two is the most important contributor to overall performance. As for the registration performance, the shape term seems to be the most important, but all terms contribute to the overall performance of the method. The positive contribution of both appearance and shape terms also confirms the hypothesis set in Section~\ref{sec:approach:stereo}. Besides, interestingly, when our model is initialized in only one of the two modalities using approximative masks, its segmentation performance is still at least as good as GrabCut's (as reported in Table~\ref{tab:segm:comp}). This highlights the robustness of our approach, and shows that it can perform well even in adverse initialization conditions.

\begin{table}[!t]
    \centering
    \rowcolors{2}{white}{gray!15}
    \begin{threeparttable}
        \renewcommand{\arraystretch}{1.35}
        \caption{Overall performance for various configurations of the proposed method on the datasets of~\cite{palmero2016,bilodeau2014}.}
        \begin{tabular}{| m{4.7cm} || M{1.0cm} M{1.0cm} |}
            \hline
            \bf{Method Configuration} & $\bar{d}_{\text{err}}$ & \multirow{1}{*}[-3pt]{$F_1$} \\[5pt]
            \hline
            \cellcolor{white}No Shape Term \scalebox{0.9}{$\big(E^{\text{shape}}\big)$}              & 8.71 & 0.860 \\
            \cellcolor{white}No Appearance Term \scalebox{0.9}{$\big(E^{\text{appearance}}\big)$}    & 3.69 & 0.851 \\
            \cellcolor{white}No Saliency Maps \scalebox{0.9}{$\big(\Xagr{W}\big)$}                   & 3.47 & 0.856 \\
            \cellcolor{white}No Uniqueness Term \scalebox{0.9}{$\big(E^{\text{uniqueness}}\big)$}    & 3.33 & 0.865 \\
            \cellcolor{white}No Color Term \scalebox{0.9}{$\big(E^{\text{color}}\big)$}              & 3.46 & 0.822 \\
            \cellcolor{white}No Contour Term \scalebox{0.9}{$\big(E^{\text{contour}}\big)$}          & 4.16 & 0.624 \\
            \cellcolor{white}No Temporal Term \scalebox{0.9}{$\big(E^{\text{temporal}}\big)$}        & 3.29 & 0.855 \\ 
            \cellcolor{white}No Initial LWIR Segm. Mask                                              &10.82 & 0.820 \\
            \cellcolor{white}No Initial Visible Segm. Mask                                           & 8.32 & 0.800 \\
            \hline
            \cellcolor{white}Default Configuration                                                   & 3.21 & 0.866 \\ 
            \hline
        \end{tabular}
        \label{tab:ablation_study}
    \end{threeparttable}
\end{table}

\begin{table}[!t]
    \centering
    \rowcolors{2}{white}{gray!15}
    \begin{threeparttable}
        \renewcommand{\arraystretch}{1.35}
        \caption{Overall segmentation performance for various temporal pipeline depths on the dataset of~\cite{palmero2016}.}
        \begin{tabular}{| m{3.5cm} || M{0.8cm} M{0.8cm} M{0.8cm} |}
            \hline
            \hiderowcolors
            \bf{Method Configuration} & \emph{Pr} & \emph{Re} & \multirow{1}{*}[-3pt]{$F_1$} \\[5pt]
            \showrowcolors
            \hline
            \cellcolor{white}2 Layers, Real-time          & 0.817 & 0.910 & 0.863 \\
            \cellcolor{white}3 Layers, Real-time          & 0.821 & 0.915 & 0.866 \\
            \cellcolor{white}4 Layers, Real-time          & 0.825 & 0.918 & 0.867 \\
            \cellcolor{white}5 Layers, Real-time          & 0.826 & 0.918 & 0.868 \\
            \hline
            \cellcolor{white}2 Layers, Deferred           & 0.821 & 0.914 & 0.866 \\ 
            \cellcolor{white}3 Layers, Deferred           & 0.824 & 0.920 & 0.870 \\
            \cellcolor{white}4 Layers, Deferred           & 0.827 & 0.921 & 0.870 \\
            \cellcolor{white}5 Layers, Deferred           & 0.826 & 0.919 & 0.868 \\
            \hline
            \rowcolor{white} No Temporal Term             & 0.801 & 0.919 & 0.855 \\ 
            \hline
        \end{tabular}
        \label{tab:temporal_study}
    \end{threeparttable}
\end{table}

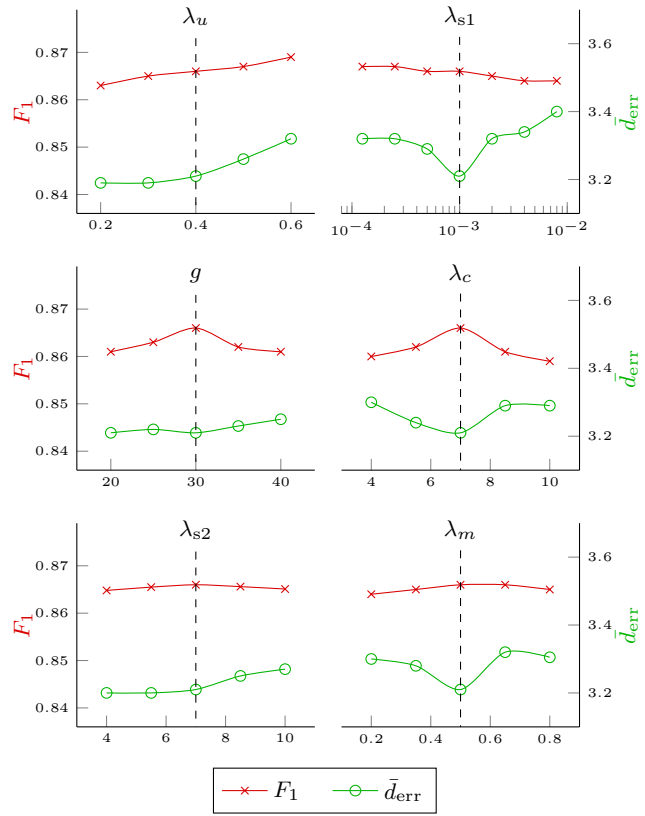
\begin{figure}[t]
\centering
\hspace{-12pt}
\begin{tikzpicture}
\begin{groupplot}[
    group style={
        group name=paramplot,
        group size=2 by 3,
        vertical sep=20pt,
        horizontal sep=10pt,
    },
    scale only axis,
    title style={yshift=-15pt,},
    width=0.18\textwidth,
    xlabel near ticks,
    legend columns=2,
    legend style={
        at={(0.4,-0.2)},
        /tikz/column 2/.style={column sep=10pt,},
        anchor=north east
    },
    ylabel shift=-5pt,
]
\newcommand\fmlowbound{0.836}
\newcommand\fmhibound{0.879}
\newcommand\adelowbound{3.10}
\newcommand\adehibound{3.70}
\pgfplotsset{every x tick label/.append style={font=\tiny, yshift=0.1ex}}
\pgfplotsset{every y tick label/.append style={font=\tiny}}
\nextgroupplot[
    title=$\lambda_u$,
    xmin=0.15, xmax=0.65, 
    ymin=\fmlowbound, ymax=\fmhibound, 
    axis y line*=left, 
    axis x line*=bottom, 
    ylabel=\color{darkred}{$F_1$},
    y tick label style={
        /pgf/number format/.cd,
            fixed,
            fixed zerofill,
            precision=2,
        /tikz/.cd
    },
]
\addplot[smooth,mark=x,darkred] coordinates{
    (0.2, 0.863)
    (0.3, 0.865)
    (0.4, 0.866) 
    (0.5, 0.867)
    (0.6, 0.869)
};
\coordinate (top_lambda_u) at (axis cs:0.4,0.873);
\coordinate (bot_lambda_u) at (axis cs:0.4,\pgfkeysvalueof{/pgfplots/ymin});
\addplot[smooth,mark=o,darkgreen,y filter/.code={\pgfmathparse{((\pgfmathresult-\adelowbound)/(\adehibound-\adelowbound))*(\fmhibound-\fmlowbound)+\fmlowbound}\pgfmathresult}] coordinates{
    (0.2, 3.19)
    (0.3, 3.19)
    (0.4, 3.21) 
    (0.5, 3.26)
    (0.6, 3.32)
};
\nextgroupplot[
    title=$\lambda_{\text{s1}}$,
    xmode=log,
    xmin=8e-5, xmax=13e-3, 
    ymin=\adelowbound, ymax=\adehibound, 
    axis y line*=right, 
    axis x line*=bottom, 
    ylabel=\color{darkgreen}{$\bar{d}_{\text{err}}$},
]
\addplot[smooth,mark=x,darkred,y filter/.code={\pgfmathparse{((\pgfmathresult-\fmlowbound)/(\fmhibound-\fmlowbound))*(\adehibound-\adelowbound)+\adelowbound}\pgfmathresult}] coordinates{
    (0.000125, 0.867)
    (0.00025, 0.867)
    (0.0005, 0.866)
    (0.001, 0.866) 
    (0.002, 0.865)
    (0.004, 0.864)
    (0.008, 0.864)
};
\coordinate (top_lambda_sone) at (axis cs:0.001,3.62);
\coordinate (bot_lambda_sone) at (axis cs:0.001,\pgfkeysvalueof{/pgfplots/ymin});
\addplot[smooth,mark=o,darkgreen] coordinates{
    (0.000125, 3.32)
    (0.00025, 3.32)
    (0.0005, 3.29)
    (0.001, 3.21) 
    (0.002, 3.32)
    (0.004, 3.34)
    (0.008, 3.40)
};
\nextgroupplot[
    title=$g$,
    xmin=16, xmax=44, 
    ymin=\fmlowbound, ymax=\fmhibound, 
    axis y line*=left, 
    axis x line*=bottom, 
    ylabel=\color{darkred}{$F_1$},
    y tick label style={
        /pgf/number format/.cd,
            fixed,
            fixed zerofill,
            precision=2,
        /tikz/.cd
    },
]
\addplot[smooth,mark=x,darkred] coordinates{
    (20, 0.861)
    (25, 0.863)
    (30, 0.866) 
    (35, 0.862)
    (40, 0.861)
};
\coordinate (top_g) at (axis cs:30,0.873);
\coordinate (bot_g) at (axis cs:30,\pgfkeysvalueof{/pgfplots/ymin});
\addplot[smooth,mark=o,darkgreen,y filter/.code={\pgfmathparse{((\pgfmathresult-\adelowbound)/(\adehibound-\adelowbound))*(\fmhibound-\fmlowbound)+\fmlowbound}\pgfmathresult}] coordinates{
    (20, 3.21)
    (25, 3.22)
    (30, 3.21) 
    (35, 3.23)
    (40, 3.25)
};
\nextgroupplot[
    title=$\lambda_c$,
    xmin=3, xmax=11, 
    ymin=\adelowbound, ymax=\adehibound, 
    axis y line*=right, 
    axis x line*=bottom, 
    ylabel=\color{darkgreen}{$\bar{d}_{\text{err}}$},
]
\addplot[smooth,mark=x,darkred,y filter/.code={\pgfmathparse{((\pgfmathresult-\fmlowbound)/(\fmhibound-\fmlowbound))*(\adehibound-\adelowbound)+\adelowbound}\pgfmathresult}] coordinates{
    (4.0, 0.860)
    (5.5, 0.862)
    (7.0, 0.866) 
    (8.5, 0.861)
    (10.0, 0.859)
};
\coordinate (top_lambda_c) at (axis cs:7.0,3.62);
\coordinate (bot_lambda_c) at (axis cs:7.0,\pgfkeysvalueof{/pgfplots/ymin});
\addplot[smooth,mark=o,darkgreen] coordinates{
    (4.0, 3.30)
    (5.5, 3.24)
    (7.0, 3.21) 
    (8.5, 3.29)
    (10.0, 3.29)
};
\nextgroupplot[
    title=$\lambda_{\text{s2}}$,
    xmin=3, xmax=11, 
    ymin=\fmlowbound, ymax=\fmhibound, 
    axis y line*=left, 
    axis x line*=bottom, 
    ylabel=\color{darkred}{$F_1$},
    y tick label style={
        /pgf/number format/.cd,
            fixed,
            fixed zerofill,
            precision=2,
        /tikz/.cd
    },
]
\addplot[smooth,mark=x,darkred] coordinates{
    (4.0, 0.8648)
    (5.5, 0.8655)
    (7.0, 0.8660) 
    (8.5, 0.8656)
    (10.0, 0.8651)
};
\coordinate (top_lambda_stwo) at (axis cs:7.0,0.873);
\coordinate (bot_lambda_stwo) at (axis cs:7.0,\pgfkeysvalueof{/pgfplots/ymin});
\addplot[smooth,mark=o,darkgreen,y filter/.code={\pgfmathparse{((\pgfmathresult-\adelowbound)/(\adehibound-\adelowbound))*(\fmhibound-\fmlowbound)+\fmlowbound}\pgfmathresult}] coordinates{
    (4.0, 3.20)
    (5.5, 3.20)
    (7.0, 3.21) 
    (8.5, 3.25)
    (10.0, 3.27)
};
\nextgroupplot[
    title=$\lambda_m$,
    xmin=0.1, xmax=0.9, 
    ymin=\adelowbound, ymax=\adehibound, 
    axis y line*=right, 
    axis x line*=bottom, 
    ylabel=\color{darkgreen}{$\bar{d}_{\text{err}}$},
]
\addplot[smooth,mark=x,darkred,y filter/.code={\pgfmathparse{((\pgfmathresult-\fmlowbound)/(\fmhibound-\fmlowbound))*(\adehibound-\adelowbound)+\adelowbound}\pgfmathresult}] coordinates{
    (0.20, 0.864)
    (0.35, 0.865)
    (0.50, 0.866) 
    (0.65, 0.866)
    (0.80, 0.865)
};
\addlegendentry{$F_1$}
\coordinate (top_lambda_m) at (axis cs:0.5,3.62);
\coordinate (bot_lambda_m) at (axis cs:0.5,\pgfkeysvalueof{/pgfplots/ymin});
\addplot[smooth,mark=o,darkgreen] coordinates{
    (0.20, 3.30)
    (0.35, 3.28)
    (0.50, 3.21) 
    (0.65, 3.32)
    (0.80, 3.305)
};
\addlegendentry{$\bar{d}_{\text{err}}$}
\end{groupplot}
\draw[dashed,thin] (top_lambda_u) -- (bot_lambda_u);
\draw[dashed,thin] (top_lambda_sone) -- (bot_lambda_sone);
\draw[dashed,thin] (top_g) -- (bot_g);
\draw[dashed,thin] (top_lambda_c) -- (bot_lambda_c);
\draw[dashed,thin] (top_lambda_stwo) -- (bot_lambda_stwo);
\draw[dashed,thin] (top_lambda_m) -- (bot_lambda_m);
\end{tikzpicture}
\caption{Overall performance for various parameter values of the proposed method on the datasets of~\cite{palmero2016,bilodeau2014}. The default configuration of each parameter is shown with the dashed line. Remember that for $F_1$, higher is better, and for $\bar{d}_{\text{err}}$, lower is better.}
\label{fig:param_tuning}
\end{figure}

Next, we show the effect of parameter tuning. The segmentation and registration performance for our proposed method in terms of overall $F_1$ score and average disparity error ($\bar{d}_{\text{err}}$, in pixels) is presented for various configurations in Figure~\ref{fig:param_tuning}. Note that we roughly tuned our method with segmentation performance as a priority to obtain our default configuration. Nonetheless, registration performance is usually near-optimal or stable around the same parameter values. In general, we can note that the choice of parameters does not seem to drastically alter our method's performance, as both metrics fairly remain stable over large value intervals.

Finally, in Table~\ref{tab:temporal_study}, we evaluate our approach configured with different temporal pipeline depths, and while allowing deferred output or not. The notion of ``pipeline depth'' here corresponds to the number of edges in the higher order temporal terms introduced in Section~\ref{sec:approach:segm}. Deferred segmentation outputs are masks generated by our method with the added latency of the full pipeline, meaning the results are evaluated with a delay equal to the pipeline depth. These masks are thus allowed more iterations in our graphical model, and benefit from more temporal information (i.e. past and future frame data). On the other hand, the real-time segmentation outputs are the masks generated by our method for all new image pairs, provided without delay. From these results, we can note that the difference between deferred and real-time output is surprisingly small. This means that our model's temporal inertia allows it to smooth out shape variations without having to peek at future frame data, which is useful for real-time surveillance systems. Besides, the overall improvements obtained by using more than two temporal layers is marginal, as more temporally consistent results also entail that some relevant shape fragments around non-rigid objects are discarded. Finally, note that using more layers results in an important increase in computational complexity: using four layers roughly triples the time required for model inference compared to the default configuration.

\subsection{LITIV 2018 Dataset}
\label{sec:experiments:litiv2018}

\begin{table*}[!t]
    \centering
    \rowcolors{3}{gray!15}{white}
    \begin{threeparttable}
        \renewcommand{\arraystretch}{1.15}
        \caption{Evaluation results for the proposed method on our newly captured multispectral video dataset.}
        \begin{tabular}{| M{2.9cm} | M{1.25cm} || M{0.9cm} M{0.9cm} | M{0.9cm} M{0.9cm} | M{0.9cm} M{0.9cm} | M{0.9cm} M{0.7cm}  M{1.2cm} |}
            \hline
            \multirow{2}{*}[-4pt]{\scalebox{0.9}{\specialcell{\bf{Evaluation Type}\\(Method)}}} & \multirow{2}{*}[-4pt]{\scalebox{0.9}{\bf{Metric}}} & \multicolumn{2}{c|}{\multirow{2}{*}[1pt]{\scalebox{0.9}{\bf{Video 1}}}} & \multicolumn{2}{c|}{\multirow{2}{*}[1pt]{\scalebox{0.9}{\bf{Video 2}}}} & \multicolumn{2}{c|}{\multirow{2}{*}[1pt]{\scalebox{0.9}{\bf{Video 3}}}} & \multicolumn{3}{c|}{\multirow{2}{*}[1pt]{\scalebox{0.9}{\bf{Overall}}}} \\[5pt]
                     &           & \scalebox{0.9}{visible} & \scalebox{0.9}{LWIR} & \scalebox{0.9}{visible} & \scalebox{0.9}{LWIR} & \scalebox{0.9}{visible} & \scalebox{0.9}{LWIR} & \scalebox{0.9}{visible} & \scalebox{0.9}{LWIR} & \scalebox{0.9}{\bf{Average}} \\
            \hline\hline
            \cellcolor{white}    & \cellcolor{white}\emph{Pr}
                & 0.933 & 0.716 & 0.938 & 0.763 & 0.935 & 0.821 & \bf{0.935} & 0.767 & 0.851 \\ 
            \cellcolor{white}    & \cellcolor{white}\emph{Re}
                & 0.721 & 0.997 & 0.834 & 0.938 & 0.750 & 0.996 & 0.768 & \bf{0.977} & \bf{0.872} \\ 
            \cellcolor{white}\multirow{-3}{*}{\scalebox{0.8}{\specialcell{\cellcolor{white}\bf{Segmentation}\\\cellcolor{white}\citep{pawcs2016}}}} & \cellcolor{white}$F_1$
                & 0.813 & 0.834 & 0.883 & 0.841 & 0.832 & 0.900 & 0.843 & 0.858 & 0.851 \\ 
            \hline
            \cellcolor{white}    & \cellcolor{white}\emph{Pr}
                & 0.883 & 0.937 & 0.874 & 0.923 & 0.921 & 0.942 & 0.893 & \bf{0.934} & \bf{0.910} \\ 
            \cellcolor{white}    & \cellcolor{white}\emph{Re}
                & 0.776 & 0.842 & 0.818 & 0.783 & 0.850 & 0.849 & \bf{0.815} & 0.825 & 0.820 \\ 
            \cellcolor{white}\multirow{-3}{*}{\scalebox{0.8}{\specialcell{\cellcolor{white}\bf{Segmentation}\\\cellcolor{white}(Proposed)}}} & \cellcolor{white}$F_1$
                & 0.826 & 0.887 & 0.845 & 0.878 & 0.884 & 0.893 & \bf{0.852} & \bf{0.876} & \bf{0.864} \\ 
            \hline\hline
            \cellcolor{white}    & \cellcolor{white}\scalebox{0.9}{\scalebox{0.9}{\%$\,$err.$\,$\textgreater 1px}}
                & 90.6 & 88.6 & 92.1 & 90.8 & 88.5 & 87.2 & 90.4 & 88.8 & 89.6 \\ 
            \cellcolor{white}    & \cellcolor{white}\scalebox{0.9}{\scalebox{0.9}{\%$\,$err.$\,$\textgreater 2px}}
                & 85.2 & 81.9 & 86.6 & 83.7 & 81.9 & 80.2 & 84.6 & 81.9 & 83.3 \\ 
            \cellcolor{white}    & \cellcolor{white}\scalebox{0.9}{\scalebox{0.9}{\%$\,$err.$\,$\textgreater 4px}}
                & 75.5 & 71.6 & 78.6 & 74.2 & 72.3 & 70.0 & 75.5 & 71.9 & 73.7 \\ 
            \cellcolor{white}\multirow{-4}{*}{\scalebox{0.8}{\specialcell{\cellcolor{white}\bf{Registration}\\\cellcolor{white}(DASC Sliding Window)}}} & \cellcolor{white}$\bar{d}_{\text{err}}$
                &30.26 &21.90 &31.22 &29.11 &26.48 &23.34 &29.32 &24.79 &27.05 \\ 
            \hline
            \cellcolor{white}    & \cellcolor{white}\scalebox{0.9}{\scalebox{0.9}{\%$\,$err.$\,$\textgreater 1px}}
                & 75.0 & 74.5 & 76.3 & 76.5 & 68.7 & 69.9 & \bf{73.3} & \bf{73.6} & \bf{73.5} \\ 
            \cellcolor{white}    & \cellcolor{white}\scalebox{0.9}{\scalebox{0.9}{\%$\,$err.$\,$\textgreater 2px}}
                & 59.5 & 59.2 & 63.4 & 63.5 & 53.3 & 54.3 & \bf{58.7} & \bf{59.0} & \bf{58.8} \\ 
            \cellcolor{white}    & \cellcolor{white}\scalebox{0.9}{\scalebox{0.9}{\%$\,$err.$\,$\textgreater 4px}}
                & 43.8 & 43.8 & 46.8 & 47.0 & 32.0 & 32.4 & \bf{40.9} & \bf{41.1} & \bf{41.0} \\ 
            \cellcolor{white}\multirow{-4}{*}{\scalebox{0.8}{\specialcell{\cellcolor{white}\bf{Registration}\\\cellcolor{white}(Proposed)}}} & \cellcolor{white}$\bar{d}_{\text{err}}$
                &26.47 &22.12 &14.43 &14.97 & 9.00 & 9.06 & \bf{16.63} &\bf{15.38} &\bf{16.01} \\ 
            \hline
        \end{tabular}
        \label{tab:litiv2018}
    \end{threeparttable}
\end{table*}

To help others compare their work on multispectral segmentation and registration, we developed and annotated a new dataset. We recorded video sequences using a stereo pair composed of a Kinect v2 for Windows (at Full HD resolution) and a FLIR A40 LWIR camera (at QVGA resolution). The sensors were roughly aligned on a fixed baseline support (approximately 50 centimeters apart) and synchronized via software to capture frame pairs at 30 Hz. Calibration data for image rectification was obtained by capturing snapshots of a foam core checkerboard pattern heated using halogen lamps to make it visible in LWIR images. For the annotations, we simultaneously recorded depth and user segmentation masks provided by the Kinect SDK, and transformed this data into foreground-background segmentation masks, adding manual touch-ups where needed. Stereo correspondences were also manually annotated like in the work of \cite{bilodeau2014} to allow an approximate evaluation of registration performance in foreground image regions. In total, this dataset contains over 6000 frame pairs split into three videos, and its groundtruth is composed of 866 binary segmentation masks and 15182 point correspondences roughly distributed among frames with visible foreground. As for the capture conditions, we deliberately recorded sequences with both strong and weak contrast between foreground and background regions in the two image modalities. More specifically, we used two different temperature calibrations to make individuals more or less perceptible in LWIR images, we introduced some cluttered background in part of the visible images, and we had people carry and exchange objects that modify their appearance in both spectral bands. Overall, this dataset should be more challenging than already available RGB-LWIR video datasets. The fact that it also allows the simultaneous evaluation of foreground segmentation and stereo registration also makes it quite unique in the current literature.

We have made this new dataset available online along with our modified version of the VAP dataset for other authors\footnote{http://www.polymtl.ca/litiv/vid/index.php}. Our Kinect's raw data which includes depth images and mapping information is also provided for those interested in trimodal segmentation tasks.

\begin{figure*}[t]
    \centering
    \hspace{-2mm}
    \begin{tabular}{c@{\hskip 1mm}c@{\hskip 1mm}c@{\hskip 1mm}c}
        \figResultsRowWideHeader{Segmentation results of \cite{pawcs2016}}{Segmentation results of the proposed method}
        \figResultsRowHeader{RGB}{LWIR}{RGB}{LWIR}
        \figResultsRow{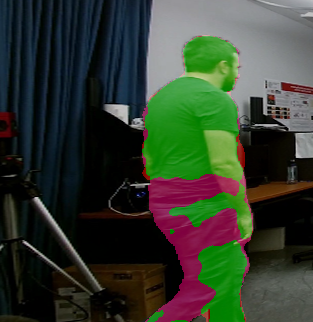}{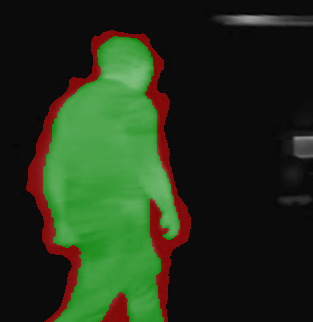}{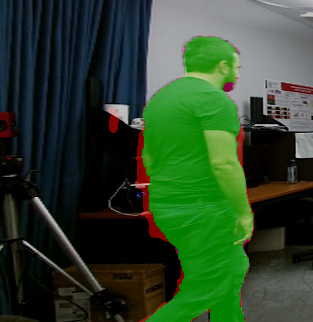}{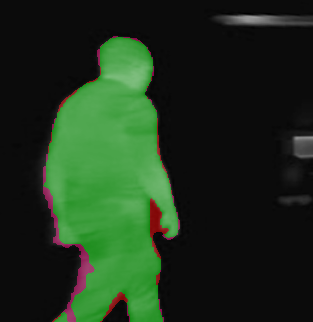}
        \figResultsRow{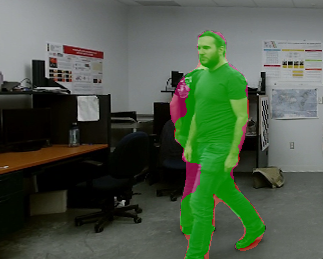}{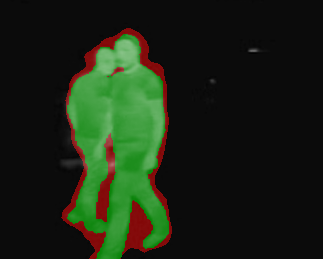}{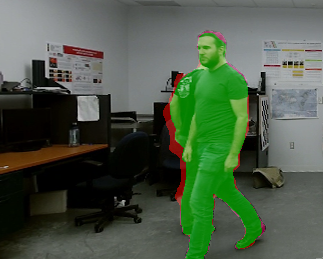}{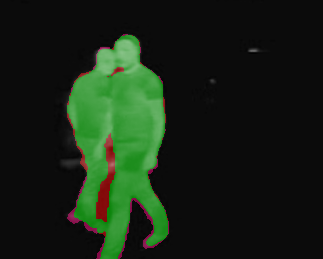}
        \figResultsRow{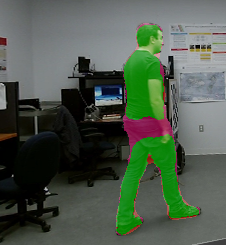}{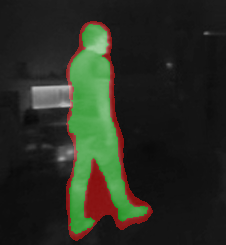}{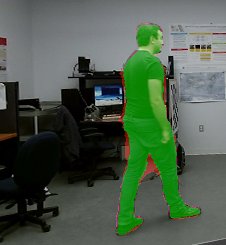}{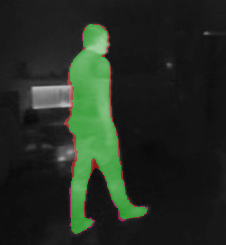}
        \figResultsRow{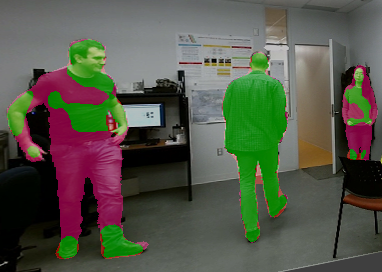}{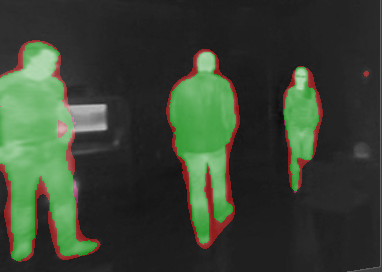}{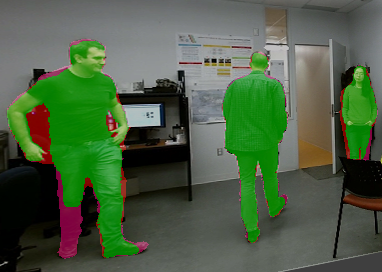}{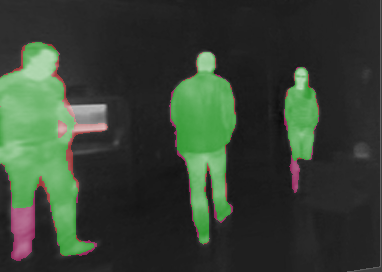}
    \end{tabular}
    \vspace{-3pt}
    \caption{Examples of typical segmentation results from our newly captured dataset; the left two columns show the segmentation masks obtained via~\cite{pawcs2016} and used to initialize our method, and the right two columns show our final segmentation masks. Image regions properly classified as foreground are highlighted in green over the original images, while regions highlighted in orange and magenta show false positives and false negatives, respectively. Images have been cropped to show more details.}
    \label{fig:res:litiv2018}
\end{figure*}

We offer our proposed method's results on this new dataset as a baseline for future comparisons in Table~\ref{tab:litiv2018}. We can note that compared to the other two datasets, segmentation results here are still good, but registration errors are much higher. This is primarily due to the fact that our camera baseline is very large ($\approx50$ cm), which leads to high disparities for close-range objects (over 150 pixels in some cases), and because our images are higher resolution than those of \cite{bilodeau2014}. Also, we can note that registration errors are higher in the first video sequence: this is caused by the loss of some small foreground segments near image borders which were annotated with correspondences. As for the segmentation results, there are cases where foreground objects are only partly detected, which results in slightly lower Recall scores in some videos. Nonetheless, these results show that our method is capable of segmentating foreground objects in difficult imaging conditions. Finally, we present qualitative segmentation results for this dataset in Figure~\ref{fig:res:litiv2018}. We can notice in the bottom row a case where segmentation errors were propagated from the visible image to the infrared one (i.e. two legs are falsely annotated as background). In short, our model can sometimes settle object boundaries in the wrong region due to occlusions in one of the views, or when strong gradients within the object happen to fit the contour model better than the object's real boundaries. A typical example of this is when a person occludes a computer monitor while wearing a shirt that is similarly colored: our model will tend to merge the monitor's contour with the person's blob. This rarely happens in practice, as a very close match in terms of visual appearance and thermal signature is required. Furthermore, as seen from the overall $F_1$ results in Table~\ref{tab:litiv2018}, our new method outperforms the previous segmentation method in both image modalities.

\section{Conclusion}
\label{sec:conclusion}

We have presented a new method for simultaneous multispectral foreground segmentation and stereo registration, and validated its capabilities on several datasets. Our approach is based on the alternating minimization of two linked energy functions that integrate multispectral shape and appearance cues. We have shown that both segmentation masks and disparity maps can simultaneously converge to good local minima without any human supervision. Furthermore, with the help of higher order factors, we achieve strong temporal coherence in our segmentation results by linking consecutive video frames inside our graphical models. To make the comparison of methods tackling this problem easier in the future, we provide our full implementation online, as well as a newly created multispectral dataset for evaluation.

If supporting large stereo baselines is unnecessary, the method could use a stronger constraint on multispectral contour similarity to improve coherence between views. Besides, explicit occlusion handling in our stereo model would further improve overall performance on the current datasets. Our model could also be generalized to provide instance-level segmentation by using a separate foreground appearance model for each object. Finally, a three-way energy minimization solution tackling foreground segmentation, stereo registration, and optical flow could be designed based on our current inference approach.


\begin{acknowledgements}

This work was supported by NSERC, by FRQ-NT team grant No. 2014-PR-172083, and by REPARTI (Regroupement pour l'\'etude des environnements partag\'es intelligents r\'epartis) FRQ-NT strategic cluster.

We gratefully acknowledge the support of NVIDIA Corporation with the donation of a Titan X GPU used for this research. We also thank Chris Holmberg Bahnsen who provided us with the full calibration data needed to rectify the stereo pairs of the VAP trimodal segmentation dataset.

\end{acknowledgements}

\bibliographystyle{spbasic}
\bibliography{document}

\end{document}